\documentclass[twoside]{article}
\usepackage[preprint]{aistats2027}
\usepackage[round]{natbib}

\usepackage{amsmath,amssymb,amsthm,mathtools,bm}
\usepackage{microtype}
\usepackage{subcaption}
\usepackage{booktabs,tabularx,array}
\usepackage{algorithm,algpseudocode}
\usepackage{xcolor}
\usepackage[hidelinks]{hyperref}
\usepackage{url}
\usepackage{enumitem}

\newcommand{\E}{\mathbb E}
\newcommand{\PP}{\mathbb P}
\newcommand{\R}{\mathbb R}

\newcolumntype{P}[1]{>{\raggedright\arraybackslash}p{#1}}
\newcolumntype{M}[1]{>{\centering\arraybackslash}m{#1}}
\theoremstyle{plain}
\newtheorem{theorem}{Theorem}[section]
\newtheorem{lemma}[theorem]{Lemma}
\newtheorem{proposition}[theorem]{Proposition}
\newtheorem{corollary}[theorem]{Corollary}
\theoremstyle{definition}

\newtheorem{assumption}[theorem]{Assumption}
\theoremstyle{remark}

\begin{document}
\runningauthor{Anish Senapati}
\twocolumn[
\aistatstitle{Convex Regression at Scale: Geometry-Preserving Constraint Sparsification}
\aistatsauthor{Anish Senapati \And Jose Blanchet}
\aistatsaddress{Institute of Computational and Mathematical Engineering\\ Stanford University }
]

\begin{abstract}
Convex regression provides a flexible nonparametric estimation framework while incorporating appropriate shape information. However, its standard least-squares formulation has $O(n^2)$ supporting-hyperplane constraints creating a computational bottleneck as the sample size $n$ increases. We study a $K$-nearest-neighbor (KNN) localized relaxation for convex regression where we for each sample, we only maintain the $K$ nearest hyperplane constraints. Under mild regularity conditions, we establish statistical guarantees for our modified estimator in every fixed dimension $d\ge2$. In particular, for $d>4$, the population risk of our modified problem can be shown to have equivalent convergence rates as the original least square convex regression estimator of $\widetilde O(n^{-2/d})$ using neighborhoods sizes (i.e. $K=K_n$) of order $\widetilde O(n^{2(d-1)/(2d-1)})$ for general convex functions and $\widetilde O(n^{(d+1)/(d+2)})$ for strongly convex functions. To corroborate our theory, we implement iterative optimization methods to compare the prediction accuracy of the KNN relaxation with those of the full estimator across synthetic convex benchmarks. Our experiments demonstrate that KNN prediction risk approaches the full-estimator prediction risk at neighborhood sizes matching the theoretical scale.
\end{abstract}

\section{Introduction}
We study the problem of estimating an unknown convex regression function from $n$ independent, identically distributed (i.i.d) observations. Convex regression arises in a wide range of areas, including economics \citep{AitSahaliaDuarte2003, Varian84}, operations research \citep{AllonEtAl2007, K11, HannahDunson2013}, statistical learning \citep{WangNi2012, LiaoDaiKuosmanen2024}, and engineering \citep{HannahDunson2012}. 
In economics, for example, concavity restrictions arise naturally in models of utility, consumer preferences, and production technologies. \citep{Afriat1967,Varian1982,Famulari1995,BlundellBrowningCrawford2003, HanochRothschild1972,AllonEtAl2007}. In financial engineering, convexity restrictions arise in option-pricing problems \citep{AitSahaliaDuarte2003}. In operations research and dynamic programming, convexity or concavity of value functions and response surfaces often follows from structural assumptions on costs, transitions, or performance measures \citep{SmithMcCardle2002,Lim2010}. 

Early work on convex and concave regression focused primarily on the one-dimensional setting establishing consistency, rates of convergence, and limit-distribution results \citep{Hildreth1954,HansonPledger1976,Mammen1991,GroeneboomJongbloedWellner2001}. More recent work has extended both the finite-dimensional characterization and the statistical theory of least-squares convex regression to multivariate data \citep{SeijoSen2011,LimGlynn2012,Lim2014,HanWellner2016}. In parallel, substantial effort has been devoted to scalable optimization methods for solving the associated quadratic program whose solution serves as the natural convex regression estimator. This quadratic program minimizes the empirical squared error over the space of max-affine functions $n$ subgradients, with each subgradient constraint encoded by $n-1$ inequalities, thus resulting in $O(n^2)$ constraints. 
 In this paper, we consider a less studied alternative approach where, instead of aiming to solve the full constrained problem, we analyze a reduced formulation in which $o(n^2)$ constraints are retained. The central question we address is:
\begin{center}
    \textit{Under what conditions can the convex regression constraints be reduced, and by how much, while preserving the same statistical convergence rate for population risk?}
\end{center}

Our main results shows that, under reasonable assumptions, we can retain only $\widetilde O(n^{2(d-1)/(2d-1)})$ constraints per subgradient for general convex functions and $\widetilde O(n^{(d+1)/(d+2)})$ constraints per subgradient for strongly convex functions while achieving the same statistical convergence rate as the full convex regression estimator for $d > 4$. Because our reduced problem relies of only keeping localized constraints from the full problem, one can substantially decrease the computational burden without sacrificing asymptotic statistical efficiency.

\subsection{Related work}
Convex regression belongs to the broader class of shape-restricted regression methods, which impose qualitative constraints on the regression function. Standard examples include isotonic regression, where the fitted function is constrained to be monotone with respect to a given order, and convex or concave regression, where the fitted function is constrained by a global curvature condition \citep{BarlowEtAl1972,RobertsonWrightDykstra1988,GuntuboyinaSen2018}. Isotonic regression yields estimators that are typically piecewise constant, while convex regression yields a piecewise affine estimator whose fitted values can be represented through supporting hyperplanes. Foundational work on convex and concave least-squares regression goes back to Hildreth \citeyearpar{Hildreth1954} and Hanson and Pledger \citeyearpar{HansonPledger1976}, with later work developing consistency, limit theory, and finite-dimensional characterizations in both one-dimensional and multivariate settings \citep{Mammen1991,GroeneboomJongbloedWellner2001,SeijoSen2011,LimGlynn2012}.

One line of work in shape-restricted regression studies the statistical complexity of the associated least-squares estimators. For isotonic and related shape-restricted problems, risk bounds and adaptation properties have been studied by \citet{ChatterjeeGuntuboyinaSen2015, GuntuboyinaSen2018}. For convex regression, analogous questions require controlling the size of the class of convex functions. Metric entropy bounds provide one way to quantify this complexity, and such bounds play a central role in the empirical-process arguments used to derive statistical rates \citep{GuntuboyinaSen2013}. In the one-dimensional setting, Guntuboyina and Sen \citeyearpar{GuntuboyinaSen2015} prove global risk bounds and adaptation results for the convex least-squares estimator, showing that the estimator can adapt to functions that are well approximated by piecewise affine convex functions. In multiple dimensions, \cite{Lim2014} and \cite{HanWellner2016} study convergence rates, minimax benchmarks, and risk bounds for multivariate convex regression. \cite{HanWellner2016} study global minimax rates and risk
bounds for bounded multivariate convex least-squares estimators, showing that the attainable rates depend strongly on the geometry of the support. Their analysis indicates that the bounded least square estimation problem is nearly rate-optimal only in low
dimensions, motivating the use of sieved adaptive estimators in higher dimensions. Kur, Gao, Guntuboyina, and Sen \citeyearpar{KurGaoGuntuboyinaSen2024} later make the high-dimensional suboptimality of the convex LSE precise, proving that for $d\ge 5$ the least square error (LSE) has risk of order $\widetilde\Theta(n^{-2/d})$ over several standard convex function classes, while the minimax rate is $n^{-4/(d+4)}$. Such a suboptimality was only true when $d > 5$ with convex regression leading to optimal estimators for $d \leq 5$. This gap motivated the study of alternative convex estimators and more structured formulations. In this direction, Bal{\'a}zs, Gy{\"o}rgy, and Szepesv{\'a}ri \citeyearpar{BalazsGyorgySzepesvari2015} consider least-squares estimators over max-affine function classes, where the estimator is constrained to be the maximum of a finite number of affine functions. Under boundedness and Lipschitz assumptions, they obtain near-optimal minimax rates up to logarithmic factors, but provide no polynomial-time
guarantee for the required optimization over affine partitions.
In contrast, Kur and Putterman
\citeyearpar{KurPutterman2022} attain near-minimax rates in polynomial
time by combining local affine fits on candidate simplices and their estimated errors into a globally convex estimator.

The usual multivariate convex least-squares estimator admits a finite-dimensional formulation with one fitted value and one subgradient vector per observation, together with pairwise supporting-hyperplane constraints. This formulation is a convex quadratic program, but it contains $O(n^2)$ linear inequality constraints, which makes solving the problem difficult for large samples sizes \citep{SeijoSen2011,LimGlynn2012,MazumderChoudhuryIyengarSen2019}. A parallel line of work in convex regression focuses on ease of computation. \citet{LeeJohnsonMorenoKuosmanen2013} developed more efficient algorithms for convex nonparametric least squares in production analysis. Mazumder, Choudhury, Iyengar, and Sen \citeyearpar{MazumderChoudhuryIyengarSen2019} propose a scalable augmented-Lagrangian and ascent descent method of multiplers(ADMM)-based frameworks for computing the multivariate convex LSE and its variants. Lin, Sun, and Toh \citeyearpar{LinSunToh2022} develop a proximal augmented-Lagrangian method combined with semismooth Newton subproblem solvers and a practical constraint-generation strategy. Bertsimas and Mundru \citeyearpar{BertsimasMundru2021} study sparse convex regression and develop cutting-plane methods for large-scale instances. These methods improve the computational tractability of convex regression, but they are primarily designed to solve, approximate, or regularize the full convex regression problem. Tangentially, Hannah and Dunson \citeyearpar{HannahDunson2013} constructed globally convex estimators by fitting local linear models on adaptively selected partitions and then combining them into a max-affine convex function, closely aligning with the statistical work of Balazs \citeyearpar{BalazsGyorgySzepesvari2015}. 

The use of constraint localization within regression problems is a less studied phenomenon within academic literature. To the best of our knowledge, no such methods have been theoretically studied in the literature for either convex or isotonic regression. In more classical regression settings, K-nearest-neighbor regression has been used to predict new data by constructing a weighted average of the learned training outcomes of the nearest neighbors.
\section{Setup and Framework}
\label{sec:framework}
\subsection{Notation}
We use $\|\cdot\|_2$ for the Euclidean norm of a vector and
$\|\cdot\|_F$ for the Frobenius norm of a matrix. For a function $f:\Omega\to\mathbb R$,
we write $\|f\|_\infty=\sup_{x\in\Omega}|f(x)|$ and
$\|f\|_n^2=n^{-1}\sum_{i=1}^n f(X_i)^2$.
We denote its Lipschitz constant with respect to Euclidean
distance by $\operatorname{Lip}(f)$. For a scalar $a$, we use
$[a]_+=\max\{a,0\}$ and $[a]_-=\max\{-a,0\}$ for its positive
and negative parts, respectively. For positive sequences $a_n,b_n$, the notation $a_n=o(b_n)$ means that $a_n/b_n\to0$ and $a_n = O(b_n)$ implies that $a_n/b_n \to C$ for some constant $C \geq 0$.
The notation $\widetilde O(\cdot)$ suppresses logarithmic factors in $n$. 
\subsection{Setting}
We consider a setting where our samples are drawn uniformly on a convex compact body $\Omega\subset\R^d$ for $d \ge 2$ with a nonempty interior. We denote $\mathbb P$ for the uniform probability measure on $\Omega$ and denote the diameter of the convex body be denoted as $D_{\Omega}$. For values $B,L>0$, we denote by $\mathcal F_{B,L}(\Omega)$ the class of convex functions on $\Omega$ with $\|f\|_\infty\le B$ and $\operatorname{Lip}(f)\le L$. We assume the following assumptions on our setting:

\begin{assumption}
For $n > 1$, our observations satisfy
\begin{equation}\label{eq:model}
Y_i=f_0(X_i)+\xi_i \text{ for }
X_i\overset{\mathrm{iid}}\sim\mathbb P\text{ and }
f_0\in\mathcal F_{B,L}(\Omega).
\end{equation}
where the errors $\xi_1,\ldots,\xi_n$ are mutually
independent and are also independent of the $X=(X_1,\ldots,X_n)$. 
\end{assumption}
\begin{assumption}
    Each error $\xi_i$ has mean 0 and is subgaussian with variance proxy $\sigma^2$:
\begin{equation}\label{eq:subgaussian-assumption}
\E[e^{t\xi_i}]\le e^{\sigma^2t^2/2}
\text{ for } t\in\R, i\le n,
\end{equation}
\end{assumption}

For a matrix of subgradients $G=(g_1,\ldots,g_n) \in \mathbb{R}^{n \times d}$ and values $\theta = (\theta_1, \theta_2, ..., \theta_n) \in \mathbb{R}^{n}$, we denote a supporting hyperplane residual between $X_i$ and $X_j$ as $$r_{ji}(\theta,G)=\theta_i-\theta_j-g_j^\top(X_i-X_j)$$ which measures how much each value $\theta_i$ is above the hyperplane generated from point $X_j$. We collect the function bounds in $$\mathcal C=\{(\theta,G):|\theta_i|\le B,\ \|g_i\|_2\le L\ \forall i\}.$$ Given an edge set $W$, we define $\mathcal C_W$ as the subset of $\mathcal C$ satisfying $r_{ji}(\theta,G)\ge0$ for every $(j,i)\in W$. For different edge sets $W$, this leads to different feasible sets for our least square formulation. Let the full edge set be $E_F=\{(j,i):j\ne i\}$. For a value $1\le K\le n-1$, we write $\mathcal N_K(j)$ for the Euclidean $K$ nearest observations to $X_j$ and subsequently denote the KNN edge set as $E_K=\{(j,i):i\in\mathcal N_K(j)\}$. 

To mitigate the commonly known problem of blowup of gradients near the boundary of $\Omega$ present in convex regression, we introduce a regularized version of our least square objective $$\Phi_n(\theta,G)=\frac12\|\theta-Y\|_2^2+\frac{\mu_n}{2}\|G\|_F^2.$$ The full convex regression and KNN convex regression problem are subsequently defined as
\begin{equation}\label{eq:full-convex-regression-lse}
(\theta_F,G_F)=\arg\min_{\mathcal C_{E_F}}\Phi_n,(\theta_K,G_K)=\arg\min_{\mathcal C_{E_K}}\Phi_n, 
\end{equation}
respectively. For out-of-sample prediction, we use the standard max-affine predictors defined by:
\begin{equation}\label{eq:max-affine-prediction}
\widehat f_A(x)=\max_{j\le n}\{\theta_{A,j}+g_{A,j}^\top(x-X_j)\}.
\end{equation}
where $A$ could denote either the full solution and KNN solution.
The predictors in \eqref{eq:max-affine-prediction} can be seen to be globally convex. However, a key distinction between the estimators arises in out-of-sample estimation. Full constraints put every affine piece below $\theta_{F,i}$ at $X_i$, while the $i$th piece attains that value thereby ensuring that $\widehat f_F(X_i)=\theta_{F,i}$. However, an omitted KNN inequality can allow another hyperplane to rise above its assigned fitted value leading to a difference seen by:
\begin{equation}\label{eq:knn-interpolation-defect}
    \widehat f_K(X_i)-\theta_{K,i}
    =
    \max_{1\le j\le n}
    \{-r_{ji}(\theta_K,G_K)\}.
\end{equation}
For notational simplicity, we define
\begin{equation}\label{eq:reference-rate}
r_{n,d}=\begin{cases}
n^{-4/(d+4)},&2\le d<4,\\
n^{-1/2}\log n,&d=4,\\
n^{-2/d},&d>4.
\end{cases}
\end{equation}
as a reference rates for our population risk. These values are the statistical rates of convergence of the full convex regression problem \citep{KurGaoGuntuboyinaSen2024}. 
To ensure that the regularization does not introduce significant bias in our estimation as the sample points grows and worsen statistical convergence rates, we further require Assumption \ref{ass:regularization} below. 
\begin{assumption}
\label{ass:regularization}
The regression penalty $\mu_n > 0$ satisfies $\mu_n \le \frac{C_\mu r_{n,d}}{L^2}$ for some fixed constant $C_\mu$.
\end{assumption}
We measure the population risk of an estimator $f_A$ with respect to a function $f_0$ through $$R_{A}(f_0) = \mathbb{E}_{f_0}[\ell_{\mathbb P}^2(\widehat f_A,f_0)] = \mathbb{E}_{f_0}[\int_{\Omega} (f_A(x) - f_0(x))^2 dx].$$ For a class of functions $\mathcal{H}$, the worst case risk is subsequently defined as $\mathcal R_{A,n}(\mathcal H)=\sup_{f_0\in\mathcal H}\E_{f_0}\ell_{\mathbb P}^2(\widehat f_A,f_0)$. Here, the expectation is taken over the sampling of the points $X_i$ and the errors $\xi_i$. 

Subsequently, our goal is to identify neighborhood sizes $K_n=o(n)$
for which the KNN estimator preserves the full estimator's
population-risk upper bound $\widetilde O(r_{n,d})$.

\section{Risk Decomposition and Bounds}
\label{sec:statistical-rates} To separate the statistical estimation error from the effect of removing constraints, consider fixing a regression function $f_0$. We can subsequently decompose the difference of $\widehat f_K-f_0$ into 
    $(\widehat f_F-f_0)+(\widehat f_K-\widehat f_F)$.
Standard algebraic manipulations yields the associated risk decomposition of
\begin{equation}\label{eq:risk-decomposition}
\begin{aligned}
    \mathbb E_{f_0}\ell_{\mathbb P}^2(\widehat f_K,f_0)
    \le{}&
    2\underbrace{\mathbb E_{f_0}\ell_{\mathbb P}^2(\widehat f_F,f_0)}_{\text{(i)}}
    +
    2\underbrace{\mathbb E_{f_0}\ell_{\mathbb P}^2(\widehat f_K,\widehat f_F)}_{\text{(ii)}}.
\end{aligned}
\end{equation}
Term~(i) measures the error of the full estimator relative to the true regression function. Term~(ii) measures the discrepancy between the KNN and full predictors fitted to the same observations. We first establish a statistical rate for term~(i). Such analyses of the convex LSE typically consider the unregularized estimator under Gaussian errors, allowing the use of Gaussian projection and empirical-process arguments \citep{KurGaoGuntuboyinaSen2024,HanWellner2016}. However, our assumptions allows more general subgaussian errors and our framework introduces a regularization term for the fitted subgradients. For those reasons, the corresponding risk analysis requires additional control of both features. An analysis of the risk for the full estimator leads to the following bound: 
\begin{proposition}[Full-estimator population risk]\label{prop:full-summary}
Assume that $f_0$ and $(X_i, Y_i)$ follow the setup in Section \ref{sec:framework}. Further assume that we use the max affine solution $\widehat f_{F}(x)$ as defined in \eqref{eq:max-affine-prediction}. Then, for sufficiently large $n$,
\begin{equation}\label{eq:full-main-expanded}
\begin{aligned}
\sup_{f_0\in\mathcal F_{B,L}(\Omega)}
\E_{f_0}\ell_{\mathbb P}^2(\widehat f_F,f_0)\le &C\bigg[(\log n)\left(r_{n,d}+\mu_nL^2+\frac{\sigma^2}{n}\right)\\
+&L^2\left(\frac{\log n}{n}\right)^{2/d}\bigg].
\end{aligned}
\end{equation}
\end{proposition}
Proposition~\ref{prop:full-summary} relies on controlling how
strongly a feasible fitted-value vector can align with the
observation noise. We first analyze the unregularized
least-squares estimator over the feasible
set $C_F$. Since the true fitted values are feasible, the optimality of the least square estimator implies that a large estimation error would require a
deviation from the truth with a correspondingly large
alignment with the noise. Standard metric entropy bounds
for bounded Lipschitz convex functions and a subgaussian
chaining argument allows us to control these alignments, yielding
an empirical squared-error bound of order
$r_{n,d}+\sigma^2/n$. We then compare the regularized and
unregularized optimization problems to show that the
slope penalty changes the fitted-value vector by at most
$\mu_nL^2$ in Euclidean norm. To transfer our sample risk to population
risk, we employ the Lipschitz bound to control the population risk for
each observation's Voronoi cell resulting in the final term of $L^2 (\log n/n)^{2/d}$. We provide a complete proof
in Appendix~\ref{app:full-analysis}.

Under Assumption~\ref{ass:regularization}, the penalty contribution does not exceed the full problem regression rate of $r_{n, d}$. We therefore obtain the population-risk
upper bound of  $\widetilde O(n^{-4/(d+4)})$ for $2\le d<4$,
$\widetilde O(n^{-1/2})$ for $d=4$, and
$\widetilde O(n^{-2/d})$ for $d>4$. Our subsequent goal is to choose a value of $K$ so that term~(ii) does not exceed this upper bound while simultaneously ensuring $K$ is $o(n)$. We begin by proving such a bound over the function class of convex functions and then move to the function class of strongly convex functions. For clarity in our exposition, we display the sufficient neighborhood size for every dimension regime and the associated risk bounds are presented in Table~\ref{tab:rates-main} where logarithmic factors have been suppressed. 

\begin{table}[!htbp]
\centering
\footnotesize
\setlength{\tabcolsep}{3pt}
\renewcommand{\arraystretch}{1.35}
\caption{Sufficient $K_n$ and associated MSE risk orders with logarithmic factors being suppressed suppressed.}
\label{tab:rates-main}
\begin{tabular}{@{}llll@{}}
\toprule
Dimension & \shortstack[l]{$K_n$ (Convex)} & \shortstack[l]{$K_n$ (Strongly Convex)} & \shortstack[l]{$R_A(f_0)$}\\
\midrule
$2\le d<4$ & $n^{\frac{(d-1)(3d+4)}{(2d-1)(d+4)}}$
& $n^{1-\frac{2d}{(d+4)(d+2)}}$ & $n^{-4/(d+4)}$\\
$d=4$ & $n^{6/7}$ & $n^{5/6}$ & $n^{-1/2}$\\
$d>4$ & $n^{\frac{2(d-1)}{2d-1}}$ & $n^{\frac{d+1}{d+2}}$ & $n^{-2/d}$\\
\bottomrule
\end{tabular}
\end{table}

\subsection{Bounded Lipschitz Convex Truths }
We first look at functions $f_0$ that are convex, bounded, and Lipschitz. We summarize our results in the theorem below:
\begin{theorem}[Convex risk rate preservation]\label{thm:general-summary}
Assume that $f_0$ and $(X_i, Y_i)$ follow the setup in Section \ref{sec:framework} and we use the max affine solution $\widehat f_{F}(x)$ as defined in \eqref{eq:max-affine-prediction}. Under Assumption \ref{ass:regularization}, if we set
\begin{equation}\label{eq:general-K-main}
\begin{split}
K_n=\min\Bigl\{n-1,\Bigl\lceil C_K\,&n^{\frac{d-1}{2d-1}}\,
(\log n)^{\frac d{2d-1}}\\
&\times r_{n,d}^{-\frac{d(d-1)}{2(2d-1)}}\Bigr\rceil\Bigr\}.
\end{split}
\end{equation}
for sufficiently large $C_K$, then we have that $K_n=o(n)$ and, for sufficiently large $n$,
\begin{equation}\label{eq:general-main-risk}
\begin{split}
\sup_{f_0\in\mathcal F_{B,L}(\Omega)}\E_{f_0}\bigl[
&(\ell_{\mathbb P}^2(\widehat f_K, f_0)\bigr]
\le C r_{n,d}\log n.
\end{split}
\end{equation}
\end{theorem}
Theorem~\ref{thm:general-summary} relies on a chaining argument where we construct paths of local edges retained in both directions from the nearest neighbor construction of edges. By combining the residual inequalities along these paths,
we show that distant pairs observations satisfy the omitted
inequalities up to a controlled error. Provided that the KNN optimizer almost satisfies the residual constraints of the full solution, we then focus on adjusting it so that it satisfied all of the full constraints. In particular, by linearly mixing the KNN
fitted values and subgradient with those of a fixed quadratic
function, we can construct a comparison point that is feasible for the full problem, Through this constructed point, first-order optimality conditions and the full estimator's
empirical-error bound to control how far the KNN fitted values
can move from the full solution. Finally, we account for the
KNN interpolation defect in \eqref{eq:knn-interpolation-defect}
and use the same argument as in the full convex regression to transfer
the fitted-value comparison to population prediction risk. We give the complete argument in Appendix~B.

\subsection{Bounded Lipschitz strongly convex truths}
Having established bounds for $K_n$ in the convex case, we now turn our attention specifically to functions $f_0$ that are strongly convex with parameter $m_0$. For fixed $m_0>0$, we denote the strongly convex function class by $\mathcal F^{\mathrm{sc}}_{B,L}(\Omega;m_0)$ as the subclass of $\mathcal F_{B,L}(\Omega)$ for which $f$ is $m_0$ strongly--convex; i.e its members satisfy
\begin{equation}\label{eq:strong-convexity-definition}
\begin{split}
f((1-t)x+ty)\le{}&(1-t)f(x)+tf(y)\\
&-\tfrac{m_0}{2}t(1-t)\|x-y\|_2^2.
\end{split}
\end{equation}
\begin{assumption}
\label{ass: strong-convexity}
The true function $f_0 \in \mathcal{F}_{B, L}^{sc}(\Omega, m_0)$ for a fixed constant $m_0$.
\end{assumption}
Under this function class, we can prove a different bound on sufficient $K$ shown in Theorem \ref{thm:strong-summary}. 
\begin{theorem}[Strong convexity rate preservation]
\label{thm:strong-summary}
Assume that $f_0$ and $(X_i, Y_i)$ follow the setup in Section \ref{sec:framework} and we use the max affine solution $\widehat f_{F}(x)$ as defined in \eqref{eq:max-affine-prediction}. Under
Assumption~\ref{ass:regularization} and \ref{ass: strong-convexity}, if we set
\begin{equation}\label{eq:strong-K-main}
K_n=
\min\left\{
n-1,\,
\left\lceil
C_K n \times (r_{n,d}\log n)^{\frac{d}{2(d+2)}}
\right\rceil
\right\},
\end{equation}
where $C_K>0$ is a sufficiently large fixed constant, then we have that $K_n=o(n)$ and, for all sufficiently large $n$,
\begin{align}
\E_{f_0}\ell_{\mathbb P}^2(\widehat f_K,f_0)
&\le C r_{n,d}\log n,
\label{eq:strong-main-risk}
\end{align}
\end{theorem}
Unlike the general convex case, Theorem \ref{thm:strong-summary} instead relies on a screening
argument in which we show that, with high probability, all active constraints in the full convex regression problem are from nearby observations residual constraints. We first show that $\widehat f_F$ approximates $f_0$ everywhere on $\Omega$ with high probability. By employing strong convexity, we can then show that every constraint connecting observations beyond a shrinking radius is strictly slack at the $(\Theta_F, G_F)$. We choose $K_n$ large enough that the KNN graph retains all constraints within this radius, and therefore every active full constraint,with
high probability. We provide the complete argument in Appendix~C.

We highlight that though both Theorem~\ref{thm:general-summary} and
Theorem~\ref{thm:strong-summary} preserve the overall population-risk
upper bound $\widetilde O(r_{n,d})$, they provide different guarantees for the both the size and effect of sparsification. Specifically, Theorem \ref{thm:general-summary} allows for the estimators that produce different predictions while only enforcing $\mathbb{E}_{f_0} l^{2}_{\mathbb P} (\widehat f_K, \widehat f_F) = \widetilde O(r_{n, d})$. In particular, we only establish that the KNN estimator achieves the required statistical rate by showing that it approximates the full estimator at the same scale. On the other hand, with the additional strong convexity assumption, we demonstrate a stronger statement: the fitted values and subgradients from the KNN
and full problems are exactly equal with probability at least $1-Cr_{n,d}^2$. Since both predictors are uniformly bounded, the complement
event contributes only $O(r_{n,d}^2)$ to the expected squared discrepancy. We therefore obtain
$\mathbb E_{f_0}\ell_{\mathbb P}^2(\widehat f_K,\widehat f_F)
=O(r_{n,d}^2)=o(r_{n,d})$. So, while Theorem \ref{thm:general-summary} is only able to control the sparsification discrepancy at the statistical upper-rate scale, the strongly convex theorem makes this discrepancy asymptotically
negligible relative to the asymptotic rate $r_{n,d}$. 

In terms of neighborhood sizes, for $d < 4$, typical calculations can show that Theorem \ref{thm:general-summary} requires fewer neighbors for risk preservation while the strongly convex theorem
provides the stronger recovery guarantee at the cost of larger sufficient neighborhood size. However, for $d>4$, the assumption of strong convexity reduces the sufficient order from
$\widetilde O(n^{2(d-1)/(2d-1)})$ to
$\widetilde O(n^{(d+1)/(d+2)})$ overtaking the neighborhood size from the general convex case. As the $\mathcal{F}_{B, L}^{sc}(\Omega, m_0)$ is a subset of $\mathcal{F}_{B, L}^{sc}(\Omega, m_0)$, the upper bounds for $d < 4$ that were achieved for convex functions can be immediately applied to strongly convex functions. However, the discrepancy between the two bounds gives indication for future improvements on the risk upper bound of our constructed KNN estimator.  
\section{Numerical experiments}
\label{sec:experiments}
To validate our population risk results, we run tests on synthetic functions and data to compare the population risk of KNN convex regression with that of full convex regression. We give general details of the implementations and numerical accuracy checks in the main text and refer the readers to Appendix~D and the Appendix~E for further implementation details and experimental settings respectively.

\subsection{Synthetic Data Construction}
We sample points $X_i$ uniformly on $\Omega=[-2,2]^d$ with $d$ ranging from 2 to 20 and selected sample sizes from $n=250$ to $n=3000$,
with $30$ independent training runs per set values of $n$ and $d$. For the true function $f_0$, we use varying different functions listed below. 

\paragraph{Quadratics} Firstly, we include a quadratic and piecewise quadratic found in previous convex regression literature. \cite{BalazsGyorgySzepesvari2015}:
\begin{equation}\label{eq:synthetic-quadratics}
\begin{aligned}
f_{\rm Q}(x)&=-8+\frac4d\|x\|_2^2,\\
f_{\rm H}(x)&=-8+\frac4d\|[x]_+\|_2^2,
\end{aligned}
\end{equation}
The full quadratic is strongly convex with curvature $8/d$, while the half quadratic is only generally convex due to its. These two problems have been seen to be problems where convex regression struggles in practice, motivating their choice in our experimentation. 

\paragraph{Maximum of Affine Planes} To test genuine slope discontinuities, we also use finite-plane truths
\begin{equation}\label{eq:synthetic-max-affine}
f_{\rm MA}(x)=\max_{1\le m\le M}\{a_m^\top x+b_m\}.
\end{equation}
We varied $M$ in the set of $\{5,10,25,50\}$, with randomly generated coefficients $a_m$m and $b_m$. These coefficients held fixed across
sample sizes and replicates for each value of $d$ and $M$.
Such function instances are convex as well as not having without any curvature arising from a quadratic term like in Eq. \eqref{eq:synthetic-quadratics} motivating their inclusion into our dataset.

\paragraph{Max Affine Estimation of a Quadratic}Finally, as a further stress test, we implement a piecewise-affine approximation to a quadratic. This function was motivated by the least square error lower-bound construction of \citet{KurGaoGuntuboyinaSen2024} which demonstrated the suboptimality of convex regression. Intuitively, we expected such a hard problem in full convex regression for possibly lead to hard problems in KNN regression. Explicitly, on the cube, we use the grid interpolant
\begin{equation}\label{eq:synthetic-hard}
\begin{aligned}
f_{{\rm hard},n}(x)&=-8+\frac4d\sum_{j=1}^d q_{J_n}(x_j),\\
q_J(t)&=\max_{0\le k<J}\{(t_k+t_{k+1})t-t_kt_{k+1}\}.
\end{aligned}
\end{equation}
Here we use the grid points $t_k=-2+4k/J$. For $d>4$, we set $J_n=\max\{2,\lceil n^{1/(2d)}\rceil\}$. The truth then has $M_n=J_n^d\asymp\sqrt n$ affine regions, while its bounds remain fixed as $n$ grows. 

For all four families, we construct them so that they have the common bounds $B=8$ and $L=16/\sqrt d$. To attempt to simulate a real scenario where we do not know such bounds apriori, we set $B = 10$ and $L = 20/\sqrt d$ as conservative estimates in our simulations.

After sampling functions, we added Gaussian noise according to $Y_i=f_0(X_i)+\xi_i$, with $\xi_i$ set so that the signal to noise ratio was set to 3. After obtaining the KNN solution and full solution for our problems, we estimated the population risk 
$R(\widehat f)=\mathbb E[(\widehat f(X)-f_0(X))^2]$ for the max-affine predictor shown in \eqref{eq:max-affine-prediction} using 10,000 data points.

In order to compare to the risk of the full estimator, standard solvers such as Gurobi and MOSEK were not able to push to the limits desired to show our asymptotic convergence. To alleviate this computational burden, iterative methods such as ADMM, dual ascent, augmented Lagrangian, and constraint generation methods have been developed in previous literature. Having tested all methods, we found that a proximal augmented Lagrangian method with subproblems solved by the semismooth Newton method developed by Lin et. el \citeyearpar{LinSunToh2022} worked exceptionally well. Further details about the method can be found in Appendix \ref{app:optimization}. 

\subsection{Results and Discussion}
For each sample size $n$, dimension $d$, and problem family, we recorded the estimated population risks of the KNN and full convex regression estimators as $K$ increased. Table~\ref{tab:knn-risk-compact} reports the ratio of these risks at $n=1500$ across several problems and dimensions. For small neighborhoods, such as $K=19$, the KNN estimator typically has substantially larger risk than the full estimator, particularly in the lower- and moderate-dimensional settings. As $K$ increases, however, we observe a steady decrease in this ratio, with the KNN and full risks becoming nearly indistinguishable at sufficiently large $K$. The slowest approach to the full-estimator risk occurs for the max-affine truth, where the ratio remains noticeably above one for larger values of $K$. While this behavior cannot be explained solely by the absence of strong convexity, since the half-quadratic and hard-grid benchmarks are also not globally strongly convex, it suggests that additional structural features in $f_0$, particularly sharp discontinuous changes in active slopes, may influence how quickly the KNN relaxation risk approaches the full estimator risk.

\begin{table*}[!htbp]
\centering
\footnotesize
\setlength{\tabcolsep}{3pt}
\renewcommand{\arraystretch}{1.05}
\caption{Ratio of empirical risk for KNN convex regression solution to full convex regression solution for different problems and dimension at $n = 1500$. We can see the convergence of risk to 1 for sufficiently large $K$.}
\label{tab:knn-risk-compact}
\begin{minipage}[t]{0.49\textwidth}
\centering
\begin{tabular*}{\linewidth}{@{\extracolsep{\fill}}rrrrr@{}}
\toprule
 & \multicolumn{4}{c}{$\widehat R_K/\widehat R_F$} \\
$d$ & $K=19$ & $K=81$ & $K=348$ & $K=501$ \\
\midrule
\multicolumn{5}{l}{\textit{Full quadratic}} \\
3 & 93.6 & 1.02 & 1 & 1 \\
4 & 86 & 1.11 & 1 & 1 \\
5 & 129 & 1.24 & 1 & 1 \\
10 & 3.7 & 1.05 & 1 & 1 \\
15 & 0.816 & 0.958 & 1 & 1 \\
\addlinespace[2pt]
\multicolumn{5}{l}{\textit{Half quadratic}} \\
3 & 172 & 1.19 & 1 & 1 \\
4 & 229 & 1.36 & 1 & 1 \\
5 & 224 & 1.76 & 1.01 & 1 \\
10 & 29.9 & 1.8 & 1.02 & 1.01 \\
15 & 2.16 & 0.953 & 0.996 & 0.999 \\
\bottomrule
\end{tabular*}
\end{minipage}
\hfill
\begin{minipage}[t]{0.49\textwidth}
\centering
\begin{tabular*}{\linewidth}{@{\extracolsep{\fill}}rrrrr@{}}
\toprule
 & \multicolumn{4}{c}{$\widehat R_K/\widehat R_F$} \\
$d$ & $K=19$ & $K=81$ & $K=348$ & $K=501$ \\
\midrule
\multicolumn{5}{l}{\textit{Max-affine (M=10)}} \\
3 & 1.33e3 & 3.84 & 1.04 & 1.02 \\
4 & 485 & 3.46 & 1.03 & 1.01 \\
5 & 696 & 3.91 & 1.06 & 1.01 \\
10 & 36.3 & 2.57 & 1.07 & 1.02 \\
15 & 4.72 & 1.39 & 1.03 & 1.01 \\
\addlinespace[2pt]
\multicolumn{5}{l}{\textit{Hard grid}} \\
3 & 126 & 1.03 & 1 & 1 \\
4 & 151 & 1.11 & 1 & 1 \\
5 & 95.2 & 1.22 & 1 & 1 \\
10 & 3.96 & 1.08 & 1 & 1 \\
15 & 0.813 & 0.948 & 1 & 1 \\
\bottomrule
\end{tabular*}
\end{minipage}
\end{table*}

\begin{figure*}[htbp]
\centering

\begin{subfigure}[t]{0.47\textwidth}
    \centering
    \includegraphics[width=\textwidth]{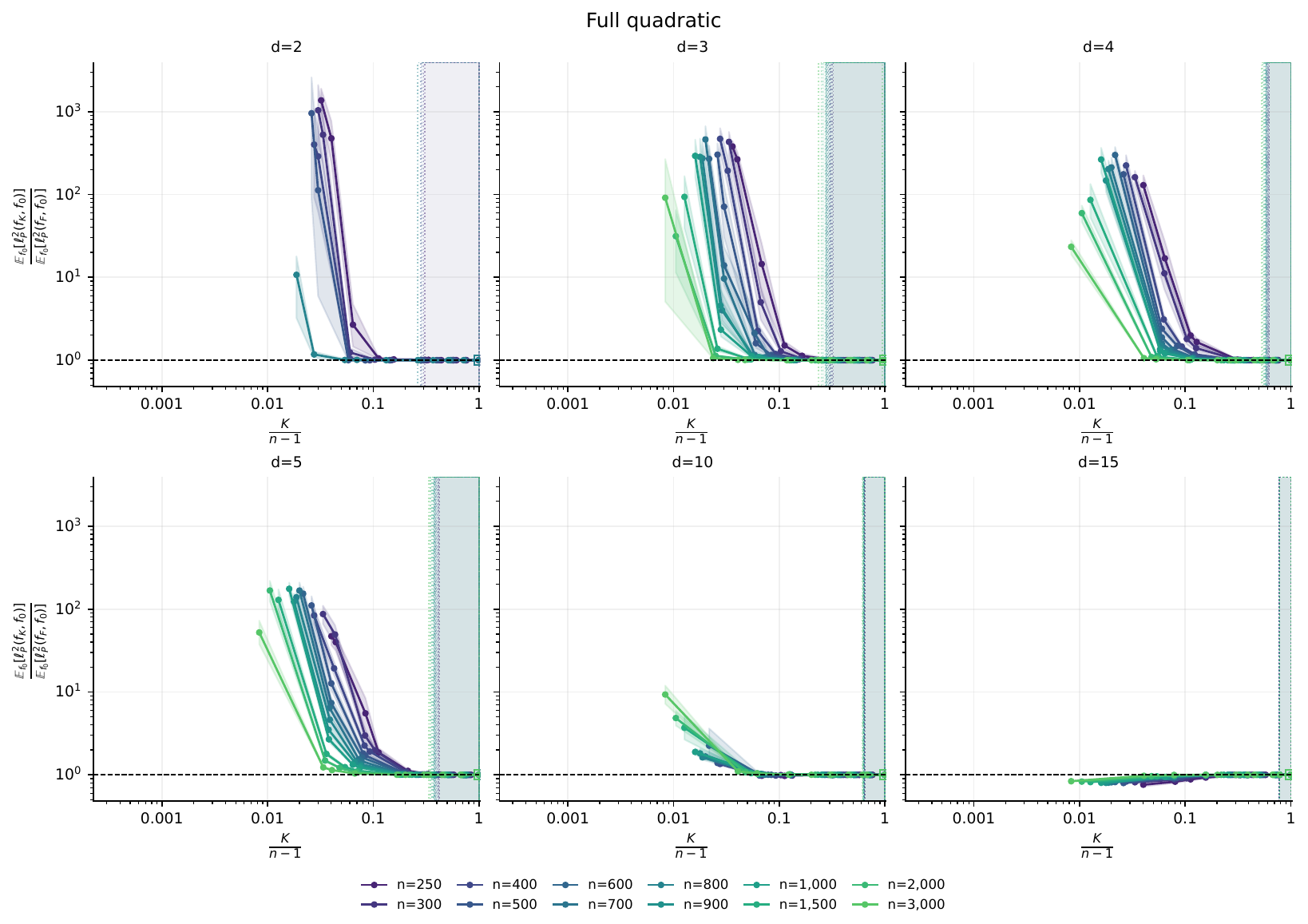}
    \caption{Full quadratic risk curve}
    \label{fig:knn-risk-1}
\end{subfigure}
\hfill
\begin{subfigure}[t]{0.47\textwidth}
    \centering
    \includegraphics[width=\textwidth]{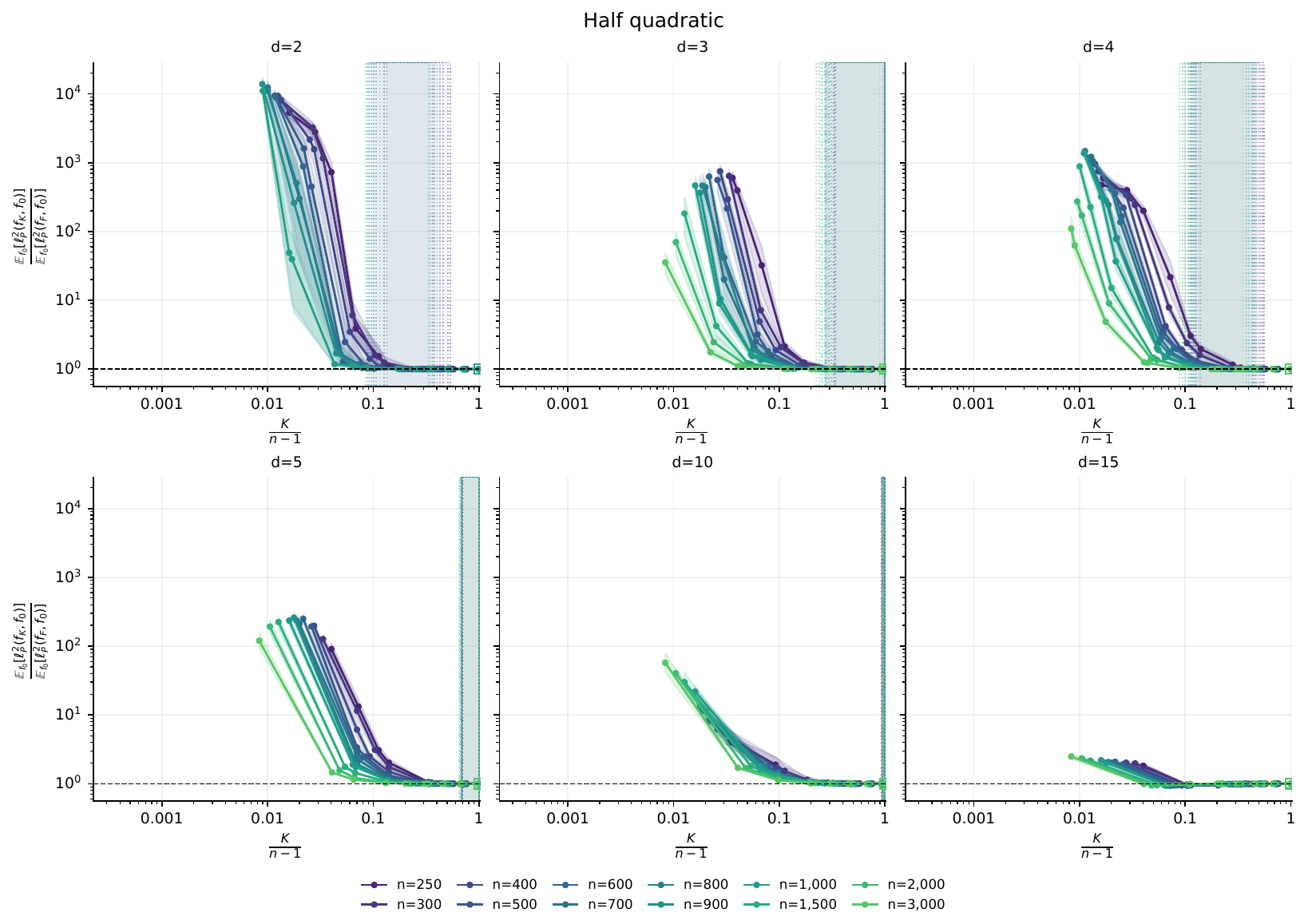}
    \caption{Half quadratic risk curve}
    \label{fig:knn-risk-2}
\end{subfigure}

\vspace{0.1cm}

\begin{subfigure}[t]{0.47\textwidth}
    \centering
    \includegraphics[width=\textwidth]{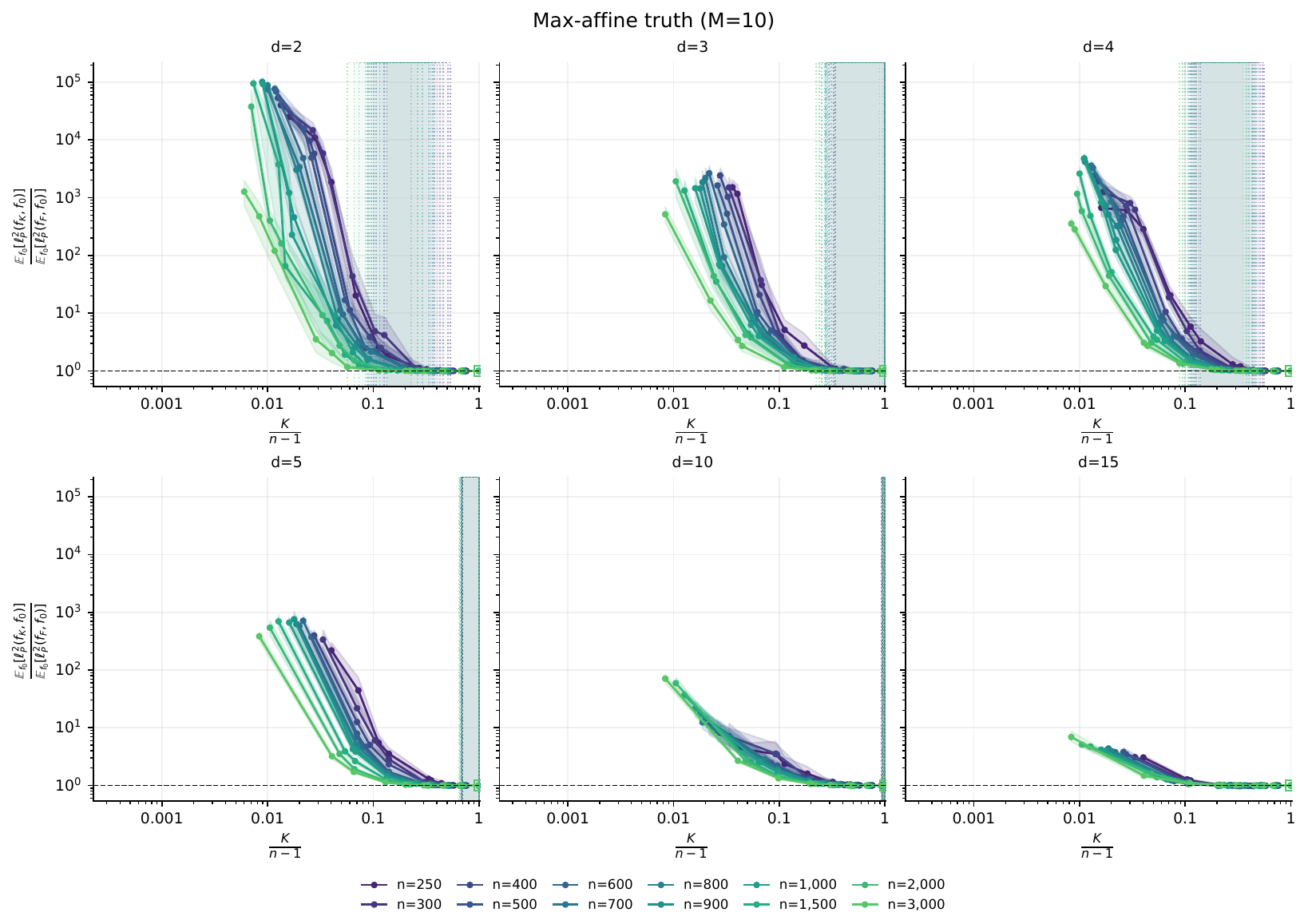}
    \caption{Max-affine truth with $M=10$.}
    \label{fig:knn-risk-3}
\end{subfigure}
\hfill
\begin{subfigure}[t]{0.47\textwidth}
    \centering
    \includegraphics[width=\textwidth]{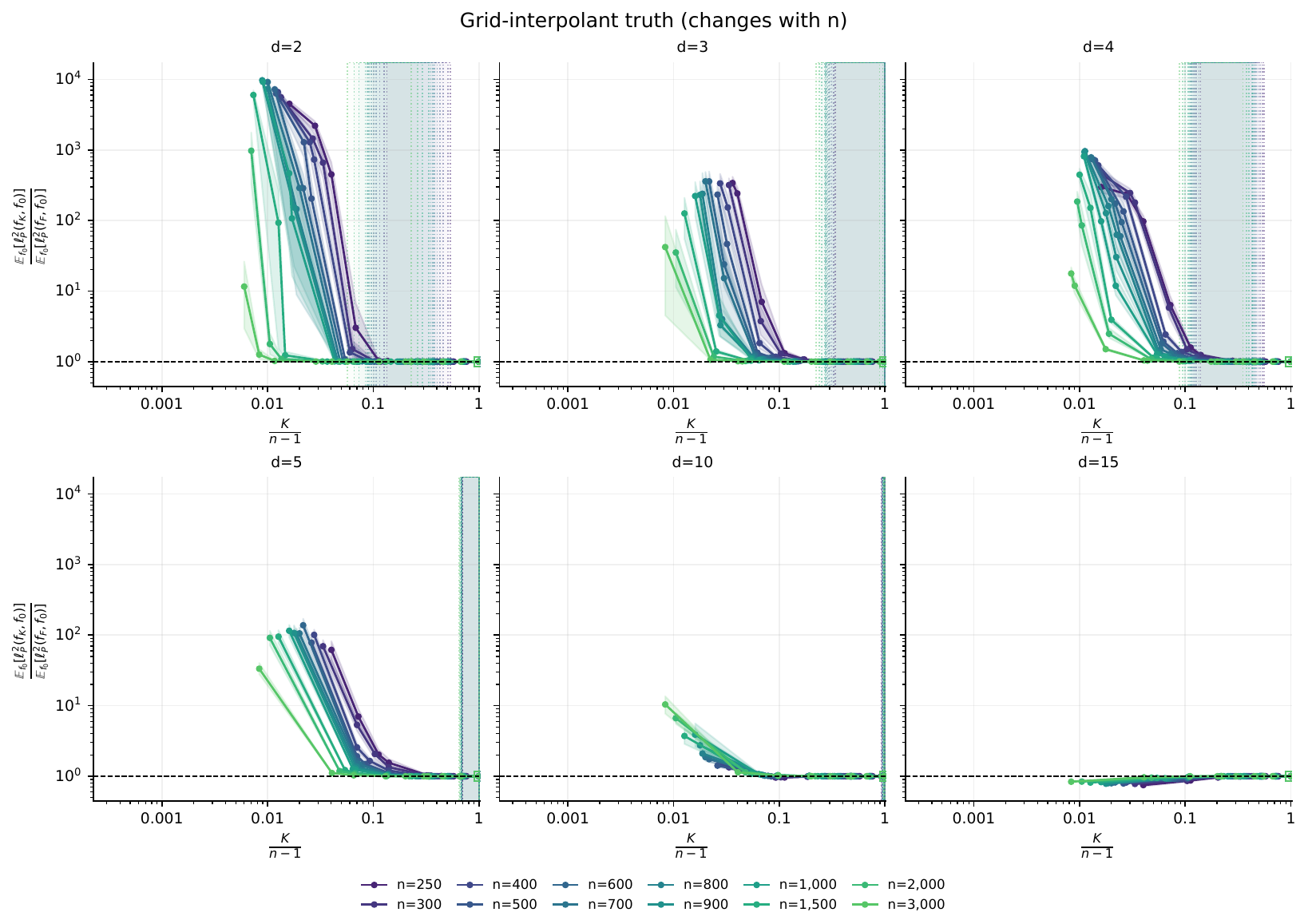}
    \caption{Grid-interpolant truth}
    \label{fig:knn-risk-4}
\end{subfigure}

\caption{Risk curves for the four synthetic convex regression truths and varying dimensions. The differing colors represent different values of $n$ and the vertical opaque bands represent what our theoretical limits mark as sufficient values of $K_n$. The risk levels converge to the full problems risk well before the predicted value of $K$.}
\label{fig:knn-risk}
\end{figure*}

As a further visualization of our experimentation, Figure~\ref{fig:knn-risk} displays the estimated population-risk ratio between the KNN solution and full solution as a function
of the fraction of retained constraints, $K/(n-1)$. Each panel corresponds to a fixed benchmark family and dimension $d$,
while the color gradient distinguishes sample sizes $n$. For a fixed value of $n$ and $d$, our theory in Theorem \ref{thm:general-summary} and \ref{thm:strong-summary} give us an estimation of the sufficient values $K_n/(n-1)$ provided large enough $n$. As such a reference rate is dependent on a multiplicative constant $C_K$, we display corresponding vertical bands extending from $0.5$ to $2$ times this reference scale. We emphasize that bands are meant to illustrate a reasonable theoretical order attained from our theory though more accurate bound would require further analyzing the dependence of $C_K$ on $d$ and the other problem parameters. We observe that the risk ratio is already close to one throughout the reference bands and generally approaches this level well before the left edges of the band corroborating our theory on the sufficiency of the values of $K_n$. In fact, we observe that we can obtain comparable prediction accuracy while retaining substantially fewer constraints than suggested by the displayed reference scales. Such a gap suggests room to sharpen our sufficient neighborhood bounds, either through accurate constant dependence or further improvements on the sample size. 

\section{Conclusion and future directions}
We introduced KNN constraint sparsification for convex regression. We showed such a sparsification preserves full-estimator risk upper bound for bounded Lipschitz convex truths under subgaussian noise. Depending on whether the true function class is strongly convex or generally convex, the level of sparsification needed to achieve the risk upper bound varies with our proof techniques. However, in both cases, a risk bound matching that of full convex regression can be achieved with a nearest neighbor sparsification $K$ that is $o(n)$. Experimental validation on sythetic data also validates such findings.

For future work, a natural next step we aim to look into would be to determine lower bound on the worst-case risk of our constructed KNN estimators and compare it with the minimax risk for appropriate function classes. Developing these lower bounds would illuminate for what type of functions constraint sparsification preserves or changes the statistical suboptimality of the full convex regression estimator.  Finally, we plan to extend our evaluation to semi-synthetic and real-world data to test the efficiency of our algorithm.

\clearpage

\section*{AI Use Statement}
In this work, we used generative AI tools for writing the code to generate our synthetic dataset, assisting in the writing of proofs, verification of our proofs, refining hypothesis, implementation of our numerical experiments, and providing feedback on our experimental design. We have not used generative AI tools for formulating our mathematical claims or providing ingredients for our proof ideas, assist with translation, clean and reformat datasets, support qualitative and thematic data analysis, or interpret results.
Additionally, we used generative AI tools for creating software code.
We have reviewed all AI-assisted work. All parts of proofs where AI were used were checked line by line by the authors and code was verified for correctness by comparing the generated methods with other standard solvers. We take responsibility for the final content of this work, including text, claims, or artifacts produced with the aid of generative AI.
\bibliography{main}

\section*{Checklist}

\begin{enumerate}

  \item For all models and algorithms presented, check if you include:
  \begin{enumerate}
    \item A clear description of the mathematical setting, assumptions, algorithm, and/or model. [\textcolor{red}{Yes}/No/Not Applicable]
    \item An analysis of the properties and complexity (time, space, sample size) of any algorithm. [\textcolor{red}{Yes}/No/Not Applicable]
    \item (Optional) Anonymized source code, with specification of all dependencies, including external libraries. [Yes/No/\textcolor{red}{Not Applicable}]
  \end{enumerate}

  \item For any theoretical claim, check if you include:
  \begin{enumerate}
    \item Statements of the full set of assumptions of all theoretical results. [\textcolor{red}{Yes}/No/Not Applicable]
    \item Complete proofs of all theoretical results. [\textcolor{red}{Yes}/No/Not Applicable]
    \item Clear explanations of any assumptions. [\textcolor{red}{Yes}/No/Not Applicable]
  \end{enumerate}

  \item For all figures and tables that present empirical results, check if you include:
  \begin{enumerate}
    \item The code, data, and instructions needed to reproduce the main experimental results (either in the supplemental material or as a URL). [\textcolor{red}{Yes}/No/Not Applicable]
    \item All the training details (e.g., data splits, hyperparameters, how they were chosen). [Yes/No/\textcolor{red}{Not Applicable}]
    \item A clear definition of the specific measure or statistics and error bars (e.g., with respect to the random seed after running experiments multiple times). [\textcolor{red}{Yes}/No/Not Applicable]
    \item A description of the computing infrastructure used. (e.g., type of GPUs, internal cluster, or cloud provider). [Yes/No/\textcolor{red}{Not Applicable}]
  \end{enumerate}

  \item If you are using existing assets (e.g., code, data, models) or curating/releasing new assets, check if you include:
  \begin{enumerate}
    \item Citations of the creator If your work uses existing assets. [Yes/No/\textcolor{red}{Not Applicable}]
    \item The license information of the assets, if applicable. [Yes/No/\textcolor{red}{Not Applicable}]
    \item New assets either in the supplemental material or as a URL, if applicable. [Yes/No/\textcolor{red}{Not Applicable}]
    \item Information about consent from data providers/curators. [Yes/No/\textcolor{red}{Not Applicable}]
    \item Discussion of sensible content if applicable, e.g., personally identifiable information or offensive content. [Yes/No/\textcolor{red}{Not Applicable}]
  \end{enumerate}

  \item If you used crowdsourcing or conducted research with human subjects, check if you include:
  \begin{enumerate}
    \item The full text of instructions given to participants and screenshots. [Yes/No/\textcolor{red}{Not Applicable}]
    \item Descriptions of potential participant risks, with links to Institutional Review Board (IRB) approvals if applicable. [Yes/No/\textcolor{red}{Not Applicable}]
    \item The estimated hourly wage paid to participants and the total amount spent on participant compensation. [Yes/No/\textcolor{red}{Not Applicable}]
  \end{enumerate}

\end{enumerate}

\clearpage
\onecolumn
\appendix

\section{Full-estimator statistical analysis}
\label{app:full-analysis}

We prove the full-estimator bound in
Proposition~\ref{prop:full-summary}. We first establish the design
estimates used to transfer sample error to population error. We then
analyze the unregularized fitted vector, control the change introduced
by the slope penalty, and obtain both an expected population-risk bound
and a high-probability bound for the screening argument in Appendix~C.
Throughout Appendices~A--C, we use the setup in
Section~\ref{sec:framework} and
Assumption~\ref{ass:regularization}. We write
$\|v\|_n^2=n^{-1}\|v\|_2^2$ for vectors in $\mathbb R^n$ and use
$\ell_{\mathbb P}^2(f,g)=\int_\Omega(f-g)^2\,d\mathbb P$.
Constants denoted by $C$ or $c$ may change from line to line and depend
only on the fixed model parameters; the constants in Appendix~C may
also depend on the fixed curvature lower bound $m_0$.

\subsection{Geometry of the uniform design}
\label{subsec:design-geometry}

The sample-to-population comparison uses how well the observations
cover $\Omega$. We denote the Voronoi cell of $X_i$ by $V_i$, and define
$\rho_n=\sup_{x\in\Omega}\min_{i\le n}\|x-X_i\|_2$ and
$p_{\max}=\max_{i\le n}\mathbb P(V_i)$.
The first quantity measures the largest gap in the design; the second
measures the largest portion of the domain assigned to one observation.

\begin{lemma}
\label{lem:uniform-design-geometry}
For independent uniform observations on the compact convex body
$\Omega\subset\mathbb R^d$, there are constants $C_\rho,C_p,C>0$
such that, for $n\ge2$ and $\delta\in(0,1)$, with probability at least
$1-\delta$,
\begin{equation}\label{eq:voronoi-geometry}
 \rho_n\le C_\rho\left(\frac{\log(n/\delta)}n\right)^{1/d},
 \qquad
 p_{\max}\le C_p\frac{\log(n/\delta)}n.
\end{equation}
Moreover, $\mathbb E\rho_n^2\le C(\log n/n)^{2/d}$.
\end{lemma}

\begin{proof}
We first establish a uniform bound on the mass of balls within the domain. We then use this bound to control the covering radius. 
Because $\Omega$ has nonempty interior, there are $x_*\in\Omega$ and $r_*>0$ such that $B(x_*,r_*)\subseteq\Omega$. For a fixed  $x\in\Omega$ and $0<r\le D_\Omega$, and setting
$t:=\frac{r}{D_\Omega+r_*}$, we have that $0<t\le1$. Every point of $(1-t)x+tB(x_*,r_*)$ is a convex combination of points in $\Omega$, so it belongs to $\Omega$. Moreover, if $y\in B(x_*,r_*)$, then
\[
    \|(1-t)x+ty-x\|_2
    =t\|y-x\|_2
    \le t\bigl(\|x_*-x\|_2+\|y-x_*\|_2\bigr)
    \le t(D_\Omega+r_*)=r.
\]
It follows that
\[
    (1-t)x+tB(x_*,r_*)\subseteq\Omega\cap B(x,r).
\]
The set on the left is a ball of radius $tr_*$. Since the design distribution is uniform,
\[
    \mathbb P(B(x,r)\cap\Omega)
    \ge \frac{v_d(tr_*)^d}{|\Omega|}
    =\frac{v_dr_*^d}{|\Omega|(D_\Omega+r_*)^d}\,r^d.
\]
where $v_d$ is the volume of a unit ball in $\mathbb{R}^d$.
For the upper bound, the volume of $B(x,r)\cap\Omega$ is no greater than the volume of the entire ball. We have therefore shown that there are constants $c_0,C_0>0$, depending only on $d$ and $\Omega$, such that
\begin{equation}\label{eq:ball-masses}
    c_0r^d
    \le\mathbb P(B(x,r)\cap\Omega)
    \le C_0r^d,
    \qquad x\in\Omega,\quad 0<r\le D_\Omega.
\end{equation}
We now control $\rho_n$. 
We begin by choosing a maximal collection $\mathcal A_r\subseteq\Omega$
whose points are separated by more than $r/2$. Maximality makes this
an $r/2$-net. The balls of radius $r/4$ centered at its points are
disjoint and lie inside $B(x_\ast,D_\Omega+r/4)$, giving
\[
    |\mathcal A_r|\,v_d(r/4)^d
    \le v_d(D_\Omega+r/4)^d
    \implies
    |\mathcal A_r|\le \left(\frac{5D_\Omega}{r}\right)^d.
\]
Suppose that every ball $B(a,r/2)$ centered at a net point $a\in\mathcal A_r$ contains at least one observation. For any $x\in\Omega$, we can first find a net point within $r/2$ of $x$ and then an observation within $r/2$ of that net point. The triangle inequality places this observation within distance $r$ of $x$. Consequently, $\rho_n\le r$ on this event. Consider the complement of that event. 
By independence of the observations and \eqref{eq:ball-masses},
\begin{align*}
    \mathbb P(\rho_n>r)
    &\le\sum_{a\in\mathcal A_r}
       \mathbb P\bigl(X_i\notin B(a,r/2)\text{ for every }i\bigr)\\
    &=\sum_{a\in\mathcal A_r}
       \bigl(1-\mathbb P(B(a,r/2)\cap\Omega)\bigr)^n\\
    &\leq \sum_{a\in\mathcal A_r} (1 - c_0 (\frac{r}{2})^d)^n \\
    &\le C_1r^{-d}\exp(-c_1nr^d),
\end{align*}
for sufficiently large constants $C_1,c_1>0$. 
We set $a=\log(n/\delta)$ and
$r_{n,\delta}=(Aa/n)^{1/d}$.
When $r_{n,\delta}\ge D_\Omega$, the bound
$\rho_n\le r_{n,\delta}$ is automatic. Otherwise, the preceding
estimate gives
\[
 \mathbb P(\rho_n>r_{n,\delta})
 \le\frac{C_1n}{Aa}\exp(-c_1Aa).
\]
Since $a\ge\log2$, we set $A = \max\{\frac{2}{c_1}, \frac{C_1}{\log 2}\}$, so that
$C_1/(A\log2)\le1$ and $c_1A\ge2$. The right-hand side is then at
most $ne^{-2a}=\delta^2/n\le\delta$.
This proves the covering-radius bound with $C_\rho=A^{1/d}$.

To control the cell masses, note that $x\in V_i$ implies $\|x-X_i\|_2\le\rho_n$. Hence
\[
    V_i\subseteq B(X_i,\rho_n)\cap\Omega.
\]
Because $\rho_n\le D_\Omega$, the upper bound in \eqref{eq:ball-masses} gives $\mathbb P(V_i)\le C_0\rho_n^d$. Therefore, on the event of $\rho_n\le r_{n,\delta}$,
\[
    p_{\max}
    \le C_0r_{n,\delta}^d
    =C_0A\frac{\log(n/\delta)}{n}.
\]
Taking $C_p=C_0A$ proves the bound on $p_{max}$ occurs with probabilty $1- \delta$.

The covering-radius bound gives $\mathbb P(\rho_n>s_n)\le n^{-3}$. Since $\rho_n\le D_\Omega$ for every sample,
\begin{align*}
    \mathbb E\rho_n^2
    &=\mathbb E\bigl[\rho_n^2\mathbf 1\{\rho_n\le s_n\}\bigr]
      +\mathbb E\bigl[\rho_n^2\mathbf 1\{\rho_n>s_n\}\bigr]\\
    &\le s_n^2+D_\Omega^2n^{-3}\\
    &\le C_\rho^2\left(\frac{4\log n}{n}\right)^{2/d}
        +D_\Omega^2n^{-3}.
\end{align*}
Since $d\ge1$, we have $2/d\le2<3$, and hence $n^{-3}$ is bounded by a constant multiple of $(\log n/n)^{2/d}$ for all $n\ge2$. The result follows for sufficiently large $C$.
\end{proof}

To determine whether
all edges below a prescribed length belong to a KNN graph, we also
develop an upper bound on the number of observations in a ball.

\begin{lemma}
\label{lem:local-counts}
For $n\ge2$, $\delta\in(0,1)$, and a deterministic radius
$R\in[0,D_\Omega]$, with probability at least $1-\delta$,
\begin{equation}\label{eq:local-count}
 \max_{j\le n}\#\{i\ne j:\|X_i-X_j\|_2\le R\}
 \le C_N\bigl(nR^d+\log(n/\delta)\bigr),
\end{equation}
where $C_N$ depends only on $d$ and $\Omega$.
\end{lemma}

\begin{proof}
We write $N_j(R)=\sum_{i\ne j}\mathbf1\{\|X_i-X_j\|_2\le R\}$.
As $j$ is fixed, the indicators are independent Bernoulli variables with common success probability $p_j = \mathbb P(B(X_j, R) \cap \Omega \leq C_0 R^d$ using \eqref{eq:ball-masses}. It follows that
\begin{align*}
    \mathbb E(e^{Z_j}\mid X_j)
    &=\bigl(1+p_j(e-1)\bigr)^{n-1}\\
    &\le\exp\bigl((n-1)p_j(e-1)\bigr)\\
    &\le\exp\bigl((e-1)C_0nR^d\bigr).
\end{align*}
Applying Markov's inequality at the threshold
$T=(e-1)C_0nR^d+\log(\frac{n}{\delta})$ therefore gives
\begin{align*}
    \mathbb P(Z_j>T\mid X_j)
    &\le e^{-T}\mathbb E(e^{Z_j}\mid X_j)\\
    &\le e^{-\log(\frac{n}{\delta})}
     =\frac{\delta}{n}.
\end{align*}
Taking a union bound over all possible $j$ gives us:
\[
    \mathbb P\left(
        \max_{j\le n}Z_j>(e-1)C_0nR^d+\log(\frac{n}{\delta})
    \right)
    \le \delta.
\] implying that the radius $R$ neighborhood around any points $X_j$ contains at most $C_N(nR^{d} + \log(n/\delta))$ holds with probability at least 1-$\delta$.

\end{proof}

The next lemma transfers error at the observations to population
error. Lipschitz continuity bounds the change in error between an
observation and the points in its Voronoi cell.

\begin{lemma}[Sample-to-population comparison]\label{lem:population_to_sample}
For any two $L$-Lipschitz functions $f,g$ on $\Omega$,
\begin{equation}\label{eq:population-to-sample}
    \ell_{\mathbb P}^2(f,g)
    \le 2p_{\max}\sum_{i=1}^n(f(X_i)-g(X_i))^2
       +8L^2\rho_n^2.
\end{equation}
\end{lemma}

\begin{proof}
Because both $f$ and $g$ are $L$-Lipschitz, their difference is $2L$-Lipschitz. For every $x\in V_i$, the definition of the Voronoi cell gives $\|x-X_i\|_2\le\rho_n$. Therefore, by setting $h = f - g$
\[
    |h(x)|
    \le |h(X_i)|+|h(x)-h(X_i)|
    \le |h(X_i)|+2L\rho_n.
\]
Squaring this inequality and using $(a+b)^2\le2a^2+2b^2$, we obtain
$h(x)^2\le2h(X_i)^2+8L^2\rho_n^2$,for $x\in V_i$.
Integration over $V_i$ yields
\[
    \int_{V_i}h(x)^2\,d\mathbb P(x)
    \le 2\mathbb P(V_i)h(X_i)^2
       +8L^2\rho_n^2\mathbb P(V_i).
\]
The cells partition $\Omega$, and their probability masses sum to one. Summing the last inequality over all cells consequently gives
\[
    \int_\Omega h(x)^2\,d\mathbb P(x)
    \le2\sum_{i=1}^n\mathbb P(V_i)h(X_i)^2+8L^2\rho_n^2.
\]
The first term is at most $2p_{\max}\sum_i h(X_i)^2$ completing the lemma.
\end{proof}

\subsection{Analyzing the statistical rate term (i)}
\label{subsec:statistical-rate-term}

We now bound the full estimator's error at the observations.
We begin with the unregularized problem so that the least-squares
optimality condition directly relates estimation error to the noise.
The reference sequence $r_{n,d}$ is defined in
\eqref{eq:reference-rate}.

\subsubsection{Feasible fitted values and the least-squares projection}

We first recall the supporting-subgradient result used to show that
the true fitted values are feasible.

\begin{theorem}[\cite{Rockafellar1970}, Theorem 23.4]
\label{thm:supporting-subgradient}
For a finite convex function $f$ on an open convex set
$U\subseteq\mathbb R^d$, every $x\in U$ admits a vector
$g\in\mathbb R^d$ such that
$f(y)\ge f(x)+g^\top(y-x)$ for all $y\in U$.
\end{theorem}

The following lemma establishes existence of the estimators and the
prediction properties needed throughout the risk analysis.

\begin{lemma}
\label{lem:fitted-value-set}
\label{lem:basic-properties}
For every design in $\Omega$ and every $W\subseteq E_F$, the set $\mathcal C_W$ is nonempty, compact, and convex. Additionally, the objective
$\Phi_n$ has a unique minimizer on this set for regularization parameter $\mu_n>0$.
The fitted-value set $\Theta_n=\{\theta\in\mathbb R^n:\exists G,
(\theta,G)\in\mathcal C_{E_F}\}$ is also nonempty, compact, and convex, and contains
the true values $\theta^0=(f_0(X_1),\ldots,f_0(X_n))$ almost surely.
Every $(\theta,G)\in\mathcal C_W$ defines a convex, $L$-Lipschitz out-of-sample predictor bounded in absolute value on $\Omega$ by
$B_f=B+LD_\Omega$. Additionally, for the out of sample predictor  $\hat{f}_{\theta_F,G_F}$ arising from full solution $(\theta_F, G_F)$, we have that
$\hat{f}_{\theta_F,G_F}(X_i)=\theta_{F, i}$ for every $i$.
\end{lemma}

\begin{proof}
We observe that $(0,0)\in\mathcal C_W$ proving nonemptiness. The linear inequalities and hard bounds make the set closed, bounded, and convex. The uniqueness of the minimizer of $\Phi_n$ comes directly from the strong convexity of the function along with the convexity of $C_W$.

The set $\Theta_n$ is the image of $\mathcal C_{E_F}$ under the continuous linear map $(\theta,G)\mapsto\theta$, so it maintains the properties of being nonempty, compact, and convex.

To show that $\theta^{0}$ is contained within $\Theta_n$, note that the observations lie on the interior of $\Omega$ with probability 1. On that event, Theorem~\ref{thm:supporting-subgradient} ensures that there exists a supporting vector $g_j^0$ at each $X_j$. Using points $X_j$ and $X_j$ ensure the hyperplane constraints are satisfied. We now must show that the bounded and Lipschitz constraints are satisfied. 
For $g_j^0\ne0$, we can look at a new point
$y=X_j+t g_j^0/\|g_j^0\|_2$ with $t>0$ small enough that
$y$ is still in $\operatorname{int}(\Omega)$.
The supporting inequality and the Lipschitz bound subsequently give us that $t\|g_j^0\|_2\le f_0(y)-f_0(X_j)\le Lt$.
Thus $\|g_j^0\|_2\le L$ even when $g_j^0=0$.
Our function class assumption gives us that
$|f_0(X_i)|\le B$ thereby proving that $(\theta^0,G^0)$ is feasible for the full problem and that $\theta_0 \in \Theta_n$.

For the out-of-sample prediction, each affine piece
$a_j(x)=\theta_j+g_j^\top(x-X_j)$ is $L$-Lipschitz and bounded by $||\theta_i||_2 + ||g_j||_2 \cdot ||x - X_j||_2 \leq B+LD_\Omega$. The maximum of these affine pieces is convex and satisfies the same bounds, since
$|f_{\theta, G}(X_i) - f_{\theta, G}(X_j)| = \max_j |a_j(x)-\max_j a_j(y)|\le\max_j|a_j(x)-a_j(y)| \leq L ||x - y||_2$.

By definition, the full solution $(\theta_F, G_F)$ has an out of sample solution $f_{F}(X_i) = \max_{j}\{\theta_{F, j} + g_{F, j}^{T}(X_i - X_j)\}$ and also satisfies the supporting hyperplane inequalities of $\theta_{F, i} \geq \theta_{F, j} + g_{F, i}^{T}(X_i - X_j)$ for all $j$. Therefore, $\hat{\theta}(X_i) \leq \theta_{F, i}$. However, $\theta_{F, i} + g_{F, i}^{T} (X_i - X_i) = \theta_{F, i}$, ensuring equality. The result follows.
\end{proof}

\subsubsection{Localization Argument}
\label{subsec:basic-inequality-localization}
Our goal within the proceeding subsections is to show that $||\hat{\theta}_{LS} - \theta^{0}||_n$ is sufficiently small with high probability. We can relate the least square objective error of $||\theta - Y||_2$ to the true objective error $||\theta^{0} - Y||_2$ by substituting $Y=\theta^0+\xi$ and expanding:
\begin{equation}\label{eq:loss-noise-decomposition}
\begin{aligned}
 &\frac1{2n}\|\theta-Y\|_2^2
       -\frac1{2n}\|\theta^0-Y\|_2^2\\
 &\quad=\frac1{2n}
   \left(\|(\theta-\theta^0)-\xi\|_2^2-\|\xi\|_2^2\right)\\
 &\quad=\frac12\|\theta-\theta^0\|_n^2
       -\frac1n\xi^\top(\theta-\theta^0).
\end{aligned}
\end{equation}
For the purpose of proving a bound, for a design
$\theta^0\in\Theta_n$, we define
$T_s=\{\theta\in\Theta_n:\|\theta-\theta^0\|_n\le s\}$ and the quantities
\begin{equation}\label{eq:subgaussian-process}
 Z_\theta=\frac1n\xi^\top(\theta-\theta^0),
 \qquad U_X(s)=\sup_{\theta\in T_s}|Z_\theta|.
\end{equation}

The next lemma shows it is enough to control $U_{X}(s)$ to prove that our estimator belongs to $T_s$.
\begin{lemma}
\label{lem:basic-localization}
For every design with $\theta^0\in\Theta_n$,
\begin{equation}\label{eq:basic-inequality}
 \frac12\|\widehat\theta_{\rm LS}-\theta^0\|_n^2
 \le\frac1n\xi^\top(\widehat\theta_{\rm LS}-\theta^0).
\end{equation}
Moreover, for every $s>0$,
\begin{equation}\label{eq:star-localization}
 \{\|\widehat\theta_{\rm LS}-\theta^0\|_n>s\}
 \subseteq\{U_X(s)>s^2/2\}.
\end{equation}
\end{lemma}

\begin{proof}
By Lemma \eqref{lem:fitted-value-set}, the truth $\theta_0$ is feasible, so
$\|\widehat\theta_{\rm LS}-Y\|_2^2\le\|\theta^0-Y\|_2^2$. We therefore have that 
\begin{align}
 &0 \leq \frac1{2n}\|\hat{\theta}_{LS}-Y\|_2^2
       -\frac1{2n}\|\theta^0-Y\|_2^2\\
 &\quad=\frac1{2n}
   \left(\|(\theta-\theta^0)-\xi\|_2^2-\|\xi\|_2^2\right)\\
 &\quad=\frac{1}{2}\|\hat{\theta}_{LS}-\theta^0\|_n^2
       -\frac1n\xi^\top(\hat{\theta}_{LS}-\theta^0).
\end{align}

For a realization with
$R=\|\widehat\theta_{\rm LS}-\theta^0\|_n>s$, consider setting
$\theta_s=\theta^0+(s/R)(\widehat\theta_{\rm LS}-\theta^0)$.
Convexity of $\Theta_n$ makes this vector feasible, and
$\|\theta_s-\theta^0\|_n=s$, so $\theta_s\in T_s$.
Calculations gives
\[
 Z_{\theta_s}
 =\frac{s}{R}Z_{\widehat\theta_{\rm LS}}
 \ge\frac{s}{R}\frac{R^2}{2}
 =\frac{sR}{2}>\frac{s^2}{2}.
\]
where the inequality is from using equation \eqref{eq:basic-inequality}. 
Since $U_X(s)\ge|Z_{\theta_s}|$, the event inclusion follows.
\end{proof}

Taking the complement of the set tells us that  $\{U_X(s)<s^2/2\} \subseteq \{\|\widehat\theta_{\rm LS}-\theta^0\|_n<s\}$. Therefore a high concentration probability bound on $U_X(s)$ would also be a high concentration bound on $||\hat{\theta}_{LS} - \theta^{0}||_n$.
\subsubsection{Entropy of the localized fitted-value set}

We now bound the number of representatives needed to approximate
$T_s$. For notation, we write $N(\varepsilon,S,\|\cdot\|)$ to denote the smallest number of $\varepsilon$ balls covering $S$ using distance metric $||\cdot||$. The centers of those balls are create an $\epsilon$ cover. We use the following covering theorem for convex functions and then evaluate its representatives at the observations.

\begin{theorem}[(\cite{GuntuboyinaSen2013}, Theorem III.2)]
\label{thm:convex-function-covering}
For a rectangle $Q=\prod_{k=1}^d[a_k,b_k]$ with $a_k<b_k$ and
fixed positive numbers $M,\Gamma_1,\ldots,\Gamma_d$, we denote by
$\mathcal C(Q;M,\Gamma)$ the class of convex functions $h:Q\to\mathbb R$
that are bounded by $M$ and
have $|h(x)-h(y)|\le\Gamma_k|x_k-y_k|$ whenever $x,y$ agree in all
coordinates other than the $k$th. Then, there are constants
$c_d,\varepsilon_d>0$, depending only on $d$, such that
\begin{equation}\label{eq:external-convex-entropy}
 \log N\bigl(\varepsilon,\mathcal C(Q;M,\Gamma),
                  \|\cdot\|_{\infty}\bigr)
 \le c_d(\frac{M+\sum_{k=1}^d\Gamma_k(b_k-a_k)}{\varepsilon})^{d/2},
 \qquad 0<\varepsilon\le\varepsilon_d(M+\sum_{k=1}^d\Gamma_k(b_k-a_k)).
\end{equation}
Here $\|h\|_{\infty,Q}=\sup_{x\in Q}|h(x)|$.
\end{theorem}

Using this theorem gives us the associated bonud below.

\begin{lemma}
\label{lem:empirical-entropy}
There are constants $C,\varepsilon_0>0$ such that for every design with
$\theta^0\in\Theta_n$ and every $s>0$,
\begin{equation}\label{eq:entropy}
 \log N(\varepsilon,T_s,\|\cdot\|_n)
 \le C\varepsilon^{-d/2},
 \qquad 0<\varepsilon\le\varepsilon_0,
\end{equation}
with covering centers chosen in $T_s$.
\end{lemma}

\begin{proof}
We choose a fixed rectangle $Q = \prod_{k=1}^d[a_k,b_k]$ such that 
$\Omega \subseteq Q$.
For each $\theta\in T_s$, there exists a matrix of subgradients $G$ that gives
a function $\hat{f}_{\theta,G}$ which interpolates $\theta$ and is
convex and $L$-Lipschitz on $Q$.
For $x\in Q$, we have that satisfies
$|\theta_j+g_j^\top(x-X_j)|\le B_Q:=B+L\operatorname{diam}(Q)$.
therefore $\hat{f}_{\theta,G}$ has the same absolute bound on $Q$.

We apply Theorem~\ref{thm:convex-function-covering} with $M=B_Q$
and $\Gamma_k=L$ at radius $\varepsilon/2$ giving us covering
functions $h_1,\ldots,h_J$ with
$\log J\le c_d(2H_Q/\varepsilon)^{d/2}$ for sufficiently small
$\varepsilon$. By evaluating the covering functions at the sample points, we can construct 
$v_j=(h_j(X_1),\ldots,h_j(X_n))$ which covers $T_s$ at radius
$\varepsilon/2$ using distance metric $||\cdot||_n$ since
\[
 \|\theta-v_j\|_n^2
 =\frac1n\sum_{i=1}^n
     |f_{\theta,G}(X_i)-h_j(X_i)|^2
 \le(\varepsilon/2)^2.
\]

For each covering ball that intersects with $T_s$, we choose a point $u_j$ in
that intersection. A vector $\theta$
covered by the original ball satisfies
$\|\theta-u_j\|_n \le \varepsilon$.
So, the points $u_j$ form an $\varepsilon$-cover of $T_s$. Taking $\varepsilon_0=\varepsilon_dH_Q$ and adjusting the constant $C$ appropriate proves the lemma.
\end{proof}
We now aim to use the entropy bound to control $U_X(s)$ through a Gaussian chaining argument. Each difference between
two successive representatives is a linear function of the noise.
We recall the following elementary properties of subgaussian matrices. 

\begin{lemma}
\label{lem:subgaussian-elementary}
Assume that $\xi$ satisfies Assumption \eqref{eq:model}. Then, for every vector $a\in\mathbb R^n$ and sample points $X$,
\begin{equation}\label{eq:linear-mgf}
 \E[e^{t a^\top\xi}\mid X]
 \le e^{\sigma^2t^2\|a\|_2^2/2},\qquad t\in\mathbb R.
\end{equation}
\begin{equation}\label{eq:linear-tail}
 \PP(|a^\top\xi|>\sigma\|a\|_2\sqrt{2t}\mid X)
 \le2e^{-t}.
\end{equation}
\begin{equation} \E[\xi_i^2\mid X]\le\sigma^2 \end{equation}
\begin{equation}\label{eq:noise-norm-tail}
 \PP(\|\xi\|_2>2\sigma\sqrt{n\log3+t}\mid X)\le e^{-t} \text{ for } t > 0.
\end{equation}
\end{lemma}

\begin{proof}
Independence of $\xi$ along with the sub-gaussianity gives
\[
 \E[e^{t a^\top\xi}\mid X]
 =\prod_i\E[e^{t a_i\xi_i}\mid X]
 \le\prod_i e^{\sigma^2t^2a_i^2/2}
 =e^{\sigma^2t^2\|a\|_2^2/2}.
\]
For $a\ne0$ and $u>0$, Markov's inequality gives
\[
 \PP(a^\top\xi>u\mid X)
 \le\inf_{t>0}\exp\{-tu+\sigma^2t^2\|a\|_2^2/2\}
 =\exp\{-u^2/(2\sigma^2\|a\|_2^2)\}.
\]
by setting $t=u/(\sigma^2\|a\|_2^2)$.
Applying the same bound to $-a$ and applying a union bound gives \eqref{eq:linear-tail}.

For the second moment bound, we note that $e^u + e^{-u} \geq 2 + u^2$ 
\[
 1+\frac{t^2}{2}\E[\xi_i^2\mid X] \leq \E[e^{t \xi_i}|X] + \E[e^{-t\xi_i}|X]
\]
implying that $\E[\xi_i^2|X] \leq \frac{2e^{\sigma^2 t^2}{2} - 1}{2t^2}$. By sending $t \Rightarrow 0$ bounds this by $\sigma^2$ as desired. 

For the norm bound, equation \eqref{eq:linear-tail} implies
$\PP(e^{\xi_i^2/(4\sigma^2)}>v\mid X)\le2v^{-2}$ for $v\ge1$.
Integration along with indepdence gives
\[
 \E[e^{\xi_i^2/(4\sigma^2)}\mid X]
 \le1+2\int_1^\infty v^{-2}\,dv=3.
  \implies \E[e^{\|\xi\|_2^2/(4\sigma^2)}\mid X]\le3^n \]
An application of Markov inequality proves the result.
\eqref{eq:noise-norm-tail}.
\end{proof}

For an increment $Z_v-Z_w=n^{-1}\xi^\top(v-w)$, 
Lemma \eqref{lem:subgaussian-elementary} gives a scale proportional to
$\sigma\|v-w\|_n/\sqrt n$.
We now bound the relevant $U_X(s)$.

\begin{lemma}
\label{lem:subgaussian-chaining}
For $0<s\le2B$, $0<\beta\le s/2$, and $t\ge0$, with
probability at least $1-2e^{-t}$,
\begin{equation}\label{eq:subgaussian-chaining-tail}
 U_X(s)\le C\sigma\left[
 \beta+\frac{J_d(s,\beta)}{\sqrt n}
 +s\sqrt{\frac{t+1}{n}}\right],
\end{equation}
where
\[
 J_d(s,\beta)=
 \begin{cases}
 s^{1-d/4},&2\le d<4,\\
 1+\log(s/\beta),&d=4,\\
 \beta^{1-d/4},&d>4.
 \end{cases}
\]
\end{lemma}

\begin{proof}
We write $N_{\rm int}$ to denote the covering numbers with centers in $T_s$.
Through Lemma~\ref{lem:empirical-entropy},
we have
$\log N_{\rm int}(u,T_s,\|\cdot\|_n)\le A_0 u^{-d/2}$
for all $0<u\le \varepsilon_0$. If $\varepsilon_0 > 2B$, we can set the inequality to $0 < 0 \leq 2B$. It not, for $u > \varepsilon_0$, we have that $\log N_{\rm int}(u,T_s,\|\cdot\|_n) \leq \log N_{\rm int}(\varepsilon_0,T_s,\|\cdot\|_n) \leq A_0 \varepsilon_0^{-d/2}.$ By choosing $A = A_0 \max\{1, (\frac{2B}{\varepsilon_0})^{d/2}\}$, we have that $\log N_{\rm int}(u,T_s,\|\cdot\|_n)$ for all $0 < u \leq 2B$.

We set successively finer radii of the form $r_j=2^{-j}s$ and the smallest integer $J$ with
$r_J\le\beta$, so that $\beta/2<r_J\le\beta$.
For each value of $j$, we take an $r_j$-cover
$\mathcal T_j$ of size $M_j$ satisfying
$\log M_j\le A r_j^{-d/2}$. We set
$\mathcal T_0=\{\theta^0\}$ and $M_0=1$.
For $\theta\in T_s$, we select a representative
$\pi_j\theta\in\mathcal T_j$ such that $||\theta - \pi_j \theta||_n \leq r_j$ and define
$\pi_0\theta=\theta^0$. Since $Z_{\theta^0}=0$,
\[
 Z_\theta
 =\sum_{j=1}^J
     (Z_{\pi_j\theta}-Z_{\pi_{j-1}\theta})
   +(Z_\theta-Z_{\pi_J\theta}).
\]
We now control the finite sum and its remainder separately.

The triangle inequality gives
$\|\pi_j\theta-\pi_{j-1}\theta\|_n\le3r_j$.
For any fixed pair $v,w$ at this distance,
Equation \eqref{eq:linear-tail} with $a=(v-w)/n$ yields
\[
 \PP\left(
 |Z_v-Z_w|>\frac{3\sigma r_j}{\sqrt n}\sqrt{2z}
 \,\middle|\,X\right)\le2e^{-z}.
\]
At level $j$, at most $M_jM_{j-1}$ pairs can be used as successive representatives to $\theta$ which tells us that 
\[\PP \left(\exists (v, w) \text{ such that } 
 |Z_v-Z_w|>\frac{3\sigma r_j}{\sqrt n}\sqrt{2z}
 \,\middle|\,X\right) \leq 2M_jM_{j-1}e^{-z}\]

By choosing $z_j=t+\log(4M_jM_{j-1})+2\log(j+1)$, we get that 
\[\PP \left(\exists (v, w) \text{ such that } 
 |Z_v-Z_w|>\frac{3\sigma r_j}{\sqrt n}\sqrt{2z}
 \,\middle|\,X\right) \leq \frac{e^{-t}}{2(j+1)^2}\]

Applying another union bound over all $j$, we have that 

\[\PP \left(\exists (j, v, w) \text{ such that } 
 |Z_v-Z_w|>\frac{3\sigma r_j}{\sqrt n}\sqrt{2z}
 \,\middle|\,X\right) \leq e^{-t}\]
where we used that $\sum_{j=1}^{J} \frac{1}{(j+1)^2} \leq \sum_{j=1}^{J} \frac{1}{j(j+1)} = 1 - \frac{1}{J+1} < 1$

Therefore, with probability $1-e{-t}$, we have that every relevant pair $Z_v - Z_w$ at every level is appropriately bounded. On such an event, we have that for all $\theta\in T_s$, consecutive pairs $(\pi_j\theta, \pi_{j-1}\theta)$
\[
 \left|\sum_{j=1}^J
       (Z_{\pi_j\theta}-Z_{\pi_{j-1}\theta})\right|
 \le\frac{3\sqrt2\sigma}{\sqrt n}\sum_{j=1}^Jr_j\sqrt{z_j}.
\]
The two cover sizes satisfy
$\log(M_jM_{j-1}) < 2A r_j^{-d/2}$, so $z_j \leq \sqrt{2}{A}r_j^{-d/2} + t + \log 4 + 2\log (j+1)$. By noting that $\sqrt{x+y} \leq \sqrt{x} + \sqrt{y}$, we get that
\[
 \sum_{j=1}^Jr_j\sqrt{z_j}
 \le C_A\sum_{j=1}^Jr_j^{1-d/4}
   +C\sum_{j=1}^Jr_j\sqrt{t+1+2\log(j+1)}.
\]

We first handle the second term. We have that 
\begin{align*}
\sum_{j=1}^Jr_j\sqrt{t+1+2\log(j+1)} &\leq s \sqrt{t + \log 4} \sum_{j=1}^{J} 2^{-j} + \sqrt{2}s \sum_{j=1}^{J} 2^{-j} \sqrt{\log(j+1)} \\
&\leq s \sqrt{t+4} + 2\sqrt{2}s  
\end{align*}
where we used that $\log(j+1) \leq j$ for $j \geq 1$. We therefore have that this can be bounded by $C\sqrt{t+1}$ for large enough constants $C$.

For the first term, we have that $\frac{\beta}{2} \leq r_J \leq \beta$. For $2 \leq d < 4$, we have that $\sum_{j=1}^Jr_j^{1-d/4} \leq \frac{s^{1-d/4}}{1-2^{1-d/4}} \leq C_d s^{1-\frac{d}{4}}$ from a geometric series sum. For $d = 4$, we have that $\sum_{j=1}^Jr_j^{1-d/4} = J \leq C (1 + \log(\frac{s}{\beta}))$. When $d = 4$, $\sum_{j=1}^Jr_j^{1-d/4} \leq \frac{r_{J}^{1-d/4}}{1-2^{1-d/4}}$. Since $r_J > \frac{\beta}{2}$, the top term simplifies to $2^{d/4-1}\beta^{1-\frac{d}{4}}$ giving us a bound of $C_d \beta^{1-d/4}$.

Hence, the telescoping sum satisfies $$\sum_{j=1}^{J}
     (Z_{\pi_j\theta}-Z_{\pi_{j-1}\theta}) \leq C \sigma \left[\frac{J_d(s, \beta)}{\sqrt{n}} + s \sqrt{\frac{t+1}{n}}\right]$$

For the remainder term, note that since $r_J < \beta$,
\[
 |Z_\theta-Z_{\pi_J\theta}|
 \le\frac1n\|\xi\|_2\|\theta-\pi_J\theta\|_2
 \le\beta\|\xi\|_n.
\]
By Lemma \eqref{lem:subgaussian-elementary}, outside an event of conditional
probability at most $e^{-t}$, this is bounded by
\[
 2\sigma\beta\sqrt{\log3+t/n}
 \le C\sigma\beta+C\sigma s\sqrt{(t+1)/n}.
\]

Combining the two events proves
the bound on $U_X(s)$.
\end{proof}

An important note in Lemma~\eqref{lem:subgaussian-chaining} is that for $2 < d < 4$, we have that sending $\beta$ to 0 does not change the failure probability allowing us to remove the first term.  
\subsection{Conversion to a high probability bound}
Lemma~\eqref{lem:basic-localization} reduces the estimation problem to
choosing $s$ for which the upper bound on $U_{X}$ is smaller than
$s^2/2$. We know appropriately choose the radii $r_{n,d}$ so that the bound from Lemma~\eqref{lem:subgaussian-chaining} is satisfied. 

\begin{proposition}
\label{prop:ls-empirical-risk}
There are constants $C,n_0>0$ such that, for $n\ge n_0$, $t\ge0$, $f_0 \in \mathcal{F}_{B,L}(\Omega)$
and for almost all s
et of points $X_1, ..., X_n$,
\begin{equation}\label{eq:ls-empirical-tail}
 \PP_{f_0}\left(
 \|\widehat\theta_{\rm LS}-\theta^0\|_n^2
 >C\left(r_{n,d}+\frac{\sigma^2(t+1)}n\right)
 \,\middle|\,X\right)\le2e^{-t}.
\end{equation}
where $r_{n,d}$ was defined in \eqref{eq:reference-rate}. Consequently,
\begin{equation}\label{eq:ls-empirical-risk}
 \E_{f_0}[\|\widehat\theta_{\rm LS}-\theta^0\|_n^2\mid X]
 \le C\left(r_{n,d}+\frac{\sigma^2}{n}\right).
\end{equation}
\end{proposition}

\begin{proof}
Using Lemma \eqref{lem:basic-localization} and \eqref{lem:subgaussian-chaining}, our goal is to choose $s$ and $\beta$ so that $$C\sigma \left[\beta + \frac{J_d(s, \beta)}{\sqrt{n}} + s\sqrt{t+1}{n} \right] < \frac{s^2}{4}$$

For a given $t\ge0$, consider the value of 
$s=(A(r_{n,d}+\sigma^2(t+1)/n))^{1/2}$, where $A$ is a
sufficiently large fixed constant.
Firstly, note that both $\theta_i$ and $f(X_i)$ are bounded by $B$, so the error
cannot exceed $s$ when $s\ge2B$. We therefore consider the case where $s<2B$.

The last term in Eq. \eqref{eq:subgaussian-chaining-tail} satisfies
\[
 \frac{C\sigma s\sqrt{(t+1)/n}}{s^2}
 \le\frac{C}{\sqrt A},
\]
since $s^2\ge A\sigma^2(t+1)/n$.
Choosing $A$ large enough makes this contribution at most $s^2/8$.

When $2\le d<4$, we remove the $\beta$ term and focus on the second term $C\sigma \frac{s^{1-d/4}}{\sqrt{n}}$. Since $s\ge \sqrt{A} r_{n, d}^{1/2} = \sqrt A\,n^{-2/(d+4)}$, we get
\[
 \frac{C\sigma n^{-1/2}s^{1-d/4}}{s^2}
 =C\sigma n^{-1/2}s^{-(1+d/4)}
 \le C\sigma A^{-(d+4)/8}.
\]

When $d=4$, we can set $\beta=n^{-1/2}$.
The lower bound $s^2\ge A n^{-1/2}\log n$ ensures
$\beta\le s/2$ for all sufficiently large $n$.
Since $s<2B$, we have
$1+\log(s/\beta)\le1+\log(2B\sqrt n)\le C_B\log n$.
Therefore,
\[
 C\sigma\left[\beta+
       \frac{1+\log(s/\beta)}{\sqrt n}\right]
 \le C'\sigma n^{-1/2}\log n
 =C'\sigma r_{n,d} \leq \frac{C'\sigma}{A} s^2.
\]

When $d>4$, we can set $\beta=n^{-2/d}$.
As $\frac{s^2}{A} > r_{n,d}$, we have that $\beta \le s/2$ for large enough $A$ and $n$, since
$s\ge\sqrt A\,r_{n,d}^{1/2}$ and $r_{n,d}\to0$.
We then get that 
\[C \sigma \left[\beta + \frac{\beta^{1-d/4}}{\sqrt{n}}\right] \leq C'\sigma n^{-2/d} \leq C' \sigma r_{n,d} \leq \frac{C' \sigma}{A} s^2\]

In all three cases, we can make this at most $s^2/8$ for sufficiently large $A$. 

Combining, we have that with probability at least $1-2e^{-t}$,
we have $U_X(s)\le s^2/4$. This proves the concentration bound. 

For the expectation, we have the objective we care about is positive giving us that
\begin{align*}
 \E[\|\widehat\theta_{\rm LS}-\theta^0\|_n^2\mid X] &= \int_{0}^{\infty} P(\|\widehat\theta_{\rm LS}-\theta^0\|_n^2 > C\left(r_{n,d}+\frac{\sigma^2(t+1)}n\right))\\
 &\le C\left(r_{n,d}+\frac{\sigma^2(t+1)}n\right)+\int_0^\infty\PP(V>C\left(r_{n,d}+\frac{\sigma^2(t+1)}n\right)+u\mid X)\,du\\
 &\le C\left(r_{n,d}+\frac{\sigma^2(t+1)}n\right)+\frac{C\sigma^2}{n}\int_0^\infty2e^{-t}\,dt
 \le C'\left(r_{n,d}+\frac{\sigma^2}{n}\right).
\end{align*}
\end{proof}

The preceding proposition concerns the unregularized fitted vector.
We now compare it with the regularized full fit. The hard slope bound
limits how much objective improvement can be obtained from reducing
the slope penalty, and the quadratic loss converts this limitation
into a bound on the movement of the fitted values.

\begin{lemma}
\label{lem:regularization-comparison}
For every realized design and response vector,
\begin{equation}\label{eq:regularization-comparison}
 \|\theta_F-\widehat\theta_{\rm LS}\|_n^2\le\mu_nL^2.
\end{equation}
\end{lemma}

\begin{proof}
Consider a subgradient matrix $G_{\rm LS}$ for
$\widehat\theta_{\rm LS}$ so that $(\widehat{\theta}_{LS}, G_{LS}) \in \mathcal{C}_{E_F}$. The bounds of the gradients give
$\|G_{\rm LS}\|_F^2\le nL^2$.
The projection optimality condition is
$(\widehat\theta_{\rm LS}-Y)^\top
 (\theta_F-\widehat\theta_{\rm LS})\ge0$.
We can then note that 
\begin{equation}\label{eq:projection-distance-objective}
 \frac12\|\theta_F-\widehat\theta_{\rm LS}\|_2^2
 \le\frac12\|\theta_F-Y\|_2^2
      -\frac12\|\widehat\theta_{\rm LS}-Y\|_2^2.
\end{equation}
Since $(\theta_F, G_{F})$ optimizes our regualarized objective functions, we have that 
\[
 \frac{\mu_n}{2}
       \bigl(\|G_{\rm LS}\|_F^2-\|G_F\|_F^2\bigr)
 \le\frac{\mu_n nL^2}{2}.
\]
Dividing by $n/2$ gives us the desired result.
\end{proof}

Combining this regularization limit with the concentration bounds of the LS solution gives the following corollary. 

\begin{corollary}[Regularized empirical error]
\label{cor:full-empirical-subgaussian}
For all sufficiently large $n$, every $\delta\in(0,1)$, and almost
every design,
\begin{equation}\label{eq:regularized-empirical-tail}
 \PP_{f_0}\left(
 \|\theta_F-\theta^0\|_n^2
 >C\left(r_{n,d}+\mu_nL^2+
                \frac{\sigma^2\log(2/\delta)}n\right)
 \,\middle|\,X\right)\le\delta,
\end{equation}
and
\begin{equation}\label{eq:full-conditional-mean}
 \E_{f_0}[\|\theta_F-\theta^0\|_n^2\mid X]
 \le C\left(r_{n,d}+\mu_nL^2+\frac{\sigma^2}{n}\right).
\end{equation}
\end{corollary}

\begin{proof}
We decompose our desired term into 
\[
 \|\theta_F-\theta^0\|_n^2
 \le2\|\theta_F-\widehat\theta_{\rm LS}\|_n^2
    +2\|\widehat\theta_{\rm LS}-\theta^0\|_n^2.
\]
Lemma \eqref{lem:regularization-comparison} bounds the first term by $\mu_nL^2$.
For the second term, an application of
Proposition \eqref{prop:ls-empirical-risk} with $t=\log(2/\delta)$ and noting that $\log(\frac{2}{\delta}) + 1 \leq \left(1 + \frac{1}{\log 2}\right) \log(\frac{2}{\delta})$, 
The expectation bound immediately follows fomr the decomposition.
\end{proof}

We now convert this to a population bound.
For the remaining appendices, we define $\delta = r_{n, d}^2$. For sufficiently large $n$, we have that $\delta_n\in(0,1)$.
Through the definition of $r_{n,d}$, it is easy to check that:
\begin{equation}\label{eq:delta-log-order}
 \log(1/\delta_n)=O(\log n),
 \qquad \log(2n/\delta_n)=O(\log n).
\end{equation}

\begin{proposition}[Full risk and population confidence bound]
\label{prop:regularized-full-risk}
For all sufficiently large $n$ and for any
$f_0\in\mathcal F_{B,L}(\Omega)$,
\begin{equation}\label{eq:full-population-mean}
 \E_{f_0}\ell_{\mathbb P}^2(\widehat f_F,f_0)
 \le C\left[(\log n)
             \left(r_{n,d}+\mu_nL^2+\frac{\sigma^2}{n}\right)
       +L^2\left(\frac{\log n}{n}\right)^{2/d}\right].
\end{equation}
For every $\delta\in(0,1)$, we have that 
\begin{equation}\label{eq:full-population-confidence}
\mathbb{P}\left(\ell_{\mathbb P}^2(\widehat f_F,f_0)
 \le C\bigg[\log\left(\frac{4n}{\delta}\right)
       \left(r_{n,d}+\mu_nL^2+
                     \frac{\sigma^2\log(4/\delta)}n\right)+L^2\left(\frac{\log(4n/\delta)}n\right)^{2/d}\bigg] \right) \leq \delta.
\end{equation}
\end{proposition}

\begin{proof}
By taking expectation over sample points $X$ in \eqref{eq:full-conditional-mean} gives us that 
\begin{equation}\label{eq:full-empirical-mean}
 \E_{f_0}\|\theta_F-\theta^0\|_n^2
 \le C\left(r_{n,d}+\mu_nL^2+\frac{\sigma^2}{n}\right),
\end{equation}
Lemma~\ref{lem:population_to_sample} gives us that 
\[
 \ell_{\mathbb P}^2(\widehat f_F,f_0)
 \le2np_{\max}\|\theta_F-\theta^0\|_n^2+8L^2\rho_n^2.
\]
We apply Lemma~\ref{lem:uniform-design-geometry} with failure
probability $\delta_n$ and denote its event by $\mathcal V_n$.
Equation~\eqref{eq:delta-log-order} yields
\begin{equation}\label{eq:geometry-at-delta-n}
 np_{\max}\le C\log n,\qquad
 \rho_n^2\le C(\log n/n)^{2/d},\qquad
 \PP(\mathcal V_n^c)\le\delta_n.
\end{equation}
Both $p_{\max}$ and $\mathcal V_n$ have no dependence on $X$, so
conditioning on $X$ and using \eqref{eq:full-conditional-mean} gives
\begin{align*}
 \E\!\left[
 \ell_{\mathbb P}^2(\widehat f_F,f_0)\right] &= \E\!\left[
 \ell_{\mathbb P}^2(\widehat f_F,f_0)\mathbf1_{\mathcal V_n}\right] + \E\!\left[
 \ell_{\mathbb P}^2(\widehat f_F,f_0)\mathbf1_{\mathcal V_n^c}\right]\\
 &\quad\le2\E\!\left[np_{\max}\mathbf1_{\mathcal V_n}
                 \E[\|\theta_F-\theta^0\|_n^2]\right]
          +8L^2\E[\rho_n^2\mathbf1_{\mathcal V_n}] + \E\!\left[
 \ell_{\mathbb P}^2(\widehat f_F,f_0)\mathbf1_{\mathcal V_n^c}\right]\\
 &\quad\le C\left[(\log n)
                   \left(r_{n,d}+\mu_nL^2+\frac{\sigma^2}{n}\right)
           +L^2(\log n/n)^{2/d}\right] + \E\!\left[
 \ell_{\mathbb P}^2(\widehat f_F,f_0)\mathbf1_{\mathcal V_n^c}\right].
\end{align*}
On $\mathcal V_n^c$, the $\ell l_{\PP}^2(\hat{f}, \hat{f}_0) \leq (B_f+B)^2$ by
Lemma~\ref{lem:basic-properties}.
The second term is therefore at most $(B_f+B)^2\delta_n$.
Since $\delta_n\le r_{n,d}$ eventually, this contribution is
absorbed into \eqref{eq:full-population-mean}.

For the confidence bound, we apply
Corollary~\ref{cor:full-empirical-subgaussian} and Lemma \ref{eq:voronoi-geometry} with failure probability of $\frac{\delta}{2}$. Intersecting both events,
\[
 \|\theta_F-\theta^0\|_n^2
 \le C\left(r_{n,d}+\mu_nL^2+
                  \frac{\sigma^2\log(4/\delta)}n\right),
 \;
 np_{\max}\le C\log(4n/\delta),
 \;
 \rho_n^2\le C\left(\frac{\log(4n/\delta)}n\right)^{2/d}.
\]
with probability $1-\delta$. Substutions prove \eqref{eq:full-population-confidence}.
\end{proof}

With Assumption \eqref{ass:regularization}, we have that 
$\mu_nL^2\le C_\mu r_{n,d}$.
In each fixed dimension, $\log n/n\lesssim r_{n,d}$, so
$r_{n,d} + \mu_nL^2 + \frac{\sigma^2 \log n}{n} \leq Cr_{n,d}$. Standard calculations show that $\left(\frac{\log n}{n}\right)^{2/d} \leq r_{n, d} \log n$.
Thus
\begin{equation}\label{eq:population-scale-simplification}
 (\log n) (r_{n,d}+\mu_nL^2+\frac{\sigma^2\log n}{n})
+L^2\left(\frac{\log n}{n}\right)^{2/d}\le Cr_{n,d}\log n,
 \qquad
 \sup_{f_0\in\mathcal F_{B,L}(\Omega)}
 \E_{f_0}\ell_{\mathbb P}^2(\widehat f_F,f_0)
 \le Cr_{n,d}\log n.
\end{equation}

\section{General convex functions}
\label{app:general-proof}

We now compare the KNN and full estimators fitted to the same sample points $X$.
The proof first shows that short KNN constraints control every omitted
inequality up to a small residual. We use these residual bounds to
construct a full-feasible comparison point, and then apply the optimality conditions of the two problems to bound the movement of
the fitted values. The final step accounts for the possibility of KNN interpolation
defect and subsequently transfers the comparison to population risk.

We retain $\delta_n=r_{n,d}^2$ and define the notation
\begin{equation}\label{eq:general-error-scales}
 \ell_n=\log\left(\frac{8n^2}{\delta_n}\right),
 \;
 v_{n,K}=\left(\frac nK\right)^{1/d}
          \left(\frac{\ell_n}{K}\right)^{1/(d-1)}.
\end{equation}
The quantity $a_n$ is the empirical-error scale from Appendix~A,
whereas intuitively $v_{n,K}$ aims to measure the error accumulated along local
paths. For fixed $d$, we have $\ell_n\asymp\log n$ and, under the
Assumption \ref{ass:regularization}, $\left(r_{n,d}+\mu_nL^2+\frac{\sigma^2}{n}\right)^{1/2},
 \le Cr_{n,d}^{1/2}$. We prove a more general bound.

\begin{theorem}
\label{thm:general-detailed}
Under the setup of Section~\ref{sec:framework} and
Assumption~\ref{ass:regularization}, there are constants
$C_0,c_0,C>0$ such that, for all sufficiently large $n$ and every
deterministic integer $K$ satisfying $C_0\ell_n\le K\le c_0n$,
\begin{equation}\label{eq:general-detailed-bound}
 \sup_{f_0\in\mathcal F_{B,L}(\Omega)}
   \E_{f_0}\ell_{\mathbb P}^2(\widehat f_K,\widehat f_F)\le C\left[
     \ell_n(v_{n,K}\left(r_{n,d}+\mu_nL^2+\frac{\sigma^2}{n}\right)^{1/2}+v_{n,K}^2)
     +\left(\frac{\ell_n}{n}\right)^{2/d}+\delta_n\right].
\end{equation}
\end{theorem}

We then appropriately choose $K_n$ so that the right-hand side is no larger
than the full-estimator benchmark.

\subsection{Directional coverage by short KNN edges}

In order for a KNN chain to be useful betwen two far points $X_i$ and $X_j$, we must intuitively satisfy two conditions: (i) its edges must be short
enough to be retained by the KNN, and (ii) each step must point approximately toward
the target. Lemma \ref{lem:basic-localization} allows us to handle the first requirement. For the second conditions, we show that a thin sector pointing toward any other
point in $\Omega$ has enough probability mass to contain an observation.

\begin{lemma}
\label{lem:sector-mass}
For $0<h\le D_\Omega$, $0<\alpha\le1/4$, and $x,y\in\Omega$
with $\|y-x\|_2>h$, we set $u=(y-x)/\|y-x\|_2$ and
\[
 S_{h,\alpha}(x,u)
 =\left\{z\in\Omega:
      h/4\le\|z-x\|_2\le h/2,\quad
      \left\|\frac{z-x}{\|z-x\|_2}-u\right\|_2\le\alpha\right\}.
\]
There is a constant $c_\Omega>0$ such that
\begin{equation}\label{eq:sector-mass}
 \mathbb P(S_{h,\alpha}(x,u))\ge c_\Omega h^d\alpha^{d-1}.
\end{equation}
\end{lemma}

\begin{proof}
We construct a cylinder contained within $S_{h, \alpha}(x, u)$ look at its volume. Consider an interior ball $B(x_\ast,r_\ast)\subseteq\Omega$.
For $t\in[h/3,5h/12]$, the point $z_t=x+tu$ lies on $[x,y]$.
By setting $\lambda=h\alpha/(64D_\Omega)$. We consider all points $w_t+v$ with $v\perp u$ and
$\|v\|_2\le\lambda r_\ast$. Each such point belongs to
$\Omega$, because
\[
 w_t+v
 =
 (1-\lambda)z_t+
 \lambda\left(x_\ast+\frac{v}{\lambda}\right).
\]
These points form a thin cylinder and it can be shown that lies in the desired sector. Since $r_\ast\le D_\Omega$,
\begin{align*}
 \|w_t+v-z_t\|_2
 &=
 \|\lambda(x_\ast-z_t)+v\|_2\\
 &\le
 \lambda\|x_\ast-z_t\|_2+\|v\|_2\\
 &\le
 \lambda(D_\Omega+r_\ast)
 \le \frac{h\alpha}{32}.
\end{align*}
Thus every cylinder point is within distance $h\alpha/32$
of the corresponding point $z_t$ on the original segment.

Because $\|z_t-x\|_2=t$, the triangle inequality gives
\[
 t-\frac{h\alpha}{32}
 \le \|w_t+v-x\|_2
 \le t+\frac{h\alpha}{32}.
\]

The interval $[h/3,5h/12]$ lies inside $[h/4,h/2]$ with
a margin of $h/12$ at each end. Since
$h\alpha/32\le h/128<h/12$, it follows that
\[
 \frac h4\le\|w_t+v-x\|_2\le\frac h2.
\]

We apply this with $a=w_t+v-x$ and $b=z_t-x=tu$.
Since $t\ge h/3$, we obtain
\[
 \left\|
 \frac{w_t+v-x}{\|w_t+v-x\|_2}-u
 \right\|_2
 \le
 \frac{2\|w_t+v-z_t\|_2}{t}
 \le
 \frac{2(h\alpha/32)}{h/3}
 =
 \frac{3\alpha}{16}
 \le\alpha.
\]
Thus the cylinder lies in $S_{h,\alpha}(x,u)$ which has length $(1-\lambda)h/12\ge h/24$ and
cross-sectional radius
$\lambda r_\ast=h\alpha r_\ast/(64D_\Omega)$.
Its volume is therefore at least a fixed positive multiple of
$h(h\alpha)^{d-1}$, giving us our result.
\end{proof}

We choose the values of 
\begin{equation}\label{eq:h-alpha}
 h_n=c_h(K/n)^{1/d},
 \; \alpha_n=C_\alpha(\ell_n/K)^{1/(d-1)}.
\end{equation} with appropriate values of $c_h$ and $C_\alpha$ to apply our result on

\begin{lemma}
\label{lemma: gnk_def}
These exists constants $c_h,C_\alpha,C_0,c_0$ so that, for
all sufficiently large $n$ and $C_0\ell_n\le K\le c_0n$,
there exists $\mathcal G_{n,K}$ with
$\mathbb P(\mathcal G_{n,K}^c)\le\delta_n$
and every edge of length at most $h_n$ is retained in $\mathcal{C}_{E_K}$, and every ordered pair $(X_j,X_i)$ with distance greater
than $h_n$ has another observation in
$S_{h_n,\alpha_n}(X_j,(X_i-X_j)/\|X_i-X_j\|_2)$.
On $\mathcal G_{n,K}$, $np_{\max}\le C\ell_n$ and $\rho_n^2\le C(\ell_n/n)^{2/d}$.
\end{lemma}

\begin{proof}
By applying Lemma~\ref{lem:local-counts} at radius $h_n$
with probability $\delta_n/4$ which gives us that $\{i\ne j:\|X_i-X_j\|_2\le h_n\}.
$ is less than
$C_N(nh_n^d+\ell_n)=C_Nc_h^dK+C_N\ell_n.$
By choosing $c_h$ so that $C_Nc_h^d\le1/2$, and choosing
$C_0\ge 2C_N \implies \ell_n \leq \frac{K}{C_0} \leq \frac{K}{2C_n}$ allowing the local count to be bounded by $K$ at every source with probability $1-\delta/4$.
Therefore every observation within distance $h_n$ belongs to that
source's $K$ nearest neighbors. Applying the same conclusion at the
other endpoint retains both directions of each such edge.

Now, suppose that we have 
$(X_j,X_i)$ with $\|X_j-X_i\|_2>h_n$.
Lemma~\ref{lem:sector-mass} bounds the probability that The remaining $n-2$ observations
miss the sector by
$\exp\left(-c_\Omega(n-2)h_n^d\alpha_n^{d-1}\right).$
Our choices for $h_n$ and $\alpha_n$ make
$nh_n^d\alpha_n^{d-1}=c_h^dC_\alpha^{d-1}\ell_n$.
Choosing $C_\alpha$ large enough that
$c_\Omega c_h^dC_\alpha^{d-1}/2\ge1$ and using that $n - 2\ge \frac{n}{2}$ for $n\ge4$, the preceding probability is then at most
$e^{-\ell_n}$. A union
bound over at most $n^2$ ordered pairs gives that the probablity any such sector is empty has probablity $n^2 e^{-\ell_n} = \frac{\delta_n}{8}$.

To apply Lemma \ref{lem:local-counts} and Lemma  \ref{lem:sector-mass}, we need to ensure $\alpha_n \leq \frac{1}{4}$ and $h_n \leq D_\Omega$. we further increase $C_0$ so that
$C_0\ge(4C_\alpha)^{d-1}$, which ensures $\alpha_n\le1/4$.
We choose $c_0$ small enough that
$h_n = c_hc_0^{1/d}\le D_\Omega$. 
These choices make all preceding estimates applicable.

Finally, Lemma~\ref{lem:uniform-design-geometry} with probability $\delta_n/4$ gives the desired remaining properties. Intersecting its event with the count and occupancy events above proves the lemma through a union bound.
\end{proof}

On the event $\mathcal G_{n,K}$, Lemma \ref{lemma: gnk_def} allows us to repeatedly take a short step pointing
toward a target observation. The next lemma controls the number of
steps and the amount by which consecutive directions turn.s

\begin{lemma}
\label{lem:knn-path}
On $\mathcal G_{n,K}$, every pair at distance
$R=\|X_i-X_j\|_2>h_n$ can be joined by a path
$x_0=X_j,x_1,\ldots,x_N=X_i$ of retained bidirectional edges.
With $s_k=\|x_{k+1}-x_k\|_2$ and
$u_k=(x_{k+1}-x_k)/s_k$, we have
\begin{equation}\label{eq:path-length-count}
 N\le C R/h_n,
 \qquad \sum_{k=0}^{N-1}s_k\le CR,
\end{equation}
with
\begin{equation}\label{eq:path-turn}
 \|u_{k+1}-u_k\|_2\le C\alpha_n,
 \qquad 0\le k\le N-2.
\end{equation}
\end{lemma}

\begin{proof}
We start at $x_0=X_j$. While the remaining distance to $X_i$ is
greater than $h_n$, we can choose an observation in the $S_{h_n, \alpha_n}(X_j, (X_i - X_j)/||X_i-X_j||_2)$ from
Lemma~\ref{lemma: gnk_def}. Once the remaining distance is at most
$h_n$, we take the final direct edge.
For a nonfinal step, we write $D_k=\|X_i-x_k\|_2$ and
$e_k=(X_i-x_k)/D_k$.
The selected edge satisfies $h_n/4\le s_k\le h_n/2$ and
$\|u_k-e_k\|_2\le\alpha_n$. Hence
\begin{align*}
 D_{k+1}
 &=\|D_ke_k-s_ku_k\|_2\\
 &\le D_k-s_k+s_k\|e_k-u_k\|_2
 \le D_k-(1-\alpha_n)s_k.
\end{align*}
implying that $D_{k} - D_{k+1} \geq (1-\alpha_n) s_k \geq \frac{3h_n}{16}$ since $\alpha_n \leq \frac{1}{4}$ and $s_k \ge h_n/4$. So, Each nonfinal step decreases the remaining distance by at least
$3h_n/16$, proving termination and $N\le CR/h_n$. Additionally, $s_k \leq \frac{D_k - D_{k+1}}{1-\alpha_n}$ implying that $$s_{N-1} + \sum_{k=1}^{N-2} s_k\leq s_{N-1} + \frac{1}{1-\alpha_n} \sum_{k=0}^{N-2} (D_k - D_{k+1}) = s_{N-1} \frac{D_0 - D_{N-1}}{1-\alpha_n} \leq s_{N-1} + \frac{R}{1-\alpha_n} \leq s_{N-1} + \frac{4}{3} R \leq \frac{7}{3} R$$

To control direction changes, we apply
the inequality $||\frac{a}{||a||_2} - \frac{b}{||b||_2}||_2 \leq 2\frac{||a-b||_2}{||b||_2}$ to
$D_ke_k-s_ku_k$ and $(D_k-s_k)e_k$.
Since $s_k<D_k/2$, we obtain
\[
 \|e_{k+1}-e_k\|_2
 \le\frac{2s_k\|u_k-e_k\|_2}{D_k-s_k}
 \le2\alpha_n.
\]
A nonfinal edge direction differs from the current target direction
by at most $\alpha_n$, and the final edge has exactly the target
direction. So, we have that 
$$\|u_{k+1}-u_k\|_2\le ||u_{k+1} - e_{k+1}||_2 + ||e_{k+1} - e_{k}||_2 + ||e_k - u_{k}||_2 \leq \alpha_n + 2\alpha_n + \alpha_n \leq 4\alpha_n$$.
Every edge has length at most $h_n$, so both of its directed
constraints are retained.
\end{proof}

As our path of chained constraints is only approximately straight, we develop a bound for the residual value of the missing constraint. 

\begin{lemma}
\label{lem:omitted-residual-control}
On $\mathcal G_{n,K}$, every $(\theta,G)\in\mathcal C_{E_K}$
satisfies, for all pairs,
\begin{equation}\label{eq:all-residual-bound}
 r_{ji}(\theta,G)\ge-\kappa_{n,K}\|X_i-X_j\|_2^2,
 \qquad
 \kappa_{n,K}=\frac{CL\alpha_n}{h_n} \le Cv_{n,K}.
\end{equation}
\end{lemma}

\begin{proof}
Retained constraints already have nonnegative residuals.
For an omitted pair, $R=\|X_i-X_j\|_2>h_n$, and we use the path
from Lemma~\ref{lem:knn-path}.
We write $\vartheta_k$ and $\gamma_k$ for the KNN fitted values and
slopes at point $x_k$. We have the point the KNN solution will maintain the residual inequality from $x_{k}$ to $x_{k+1}$ and from $x_{k+1}$ to $x_k$. Therefore, $\vartheta_{k+1} \geq \vartheta_{k} + \gamma^{T}_k(x_{k+1} - x_k)$ which implies that $$\frac{\vartheta_{k+1} - \vartheta_{k}}{||x_{k+1} - x_{k}||_2} \geq \gamma_{k}^{T} \frac{(x_{k+1}-x_{k})}{||x_{k+1} - x_{k}||_2}.$$ Defining $s_k$ and $u_k$ just as in Lemma \ref{eq:path-length-count} and also performing simular calculations between $x_{k+1}$ and $x_k$, we get
\begin{equation}\label{eq:edge-secant-sandwich}
 \gamma_k^\top u_k
 \le\frac{\vartheta_{k+1}-\vartheta_k}{s_k}
 \le\gamma_{k+1}^\top u_k.
\end{equation}
In particular, $(\gamma_{k+1}-\gamma_k)^\top u_k\ge0$.

We define $b_k=(\gamma_k-\gamma_0)^\top u_k$, so $b_0=0$.
For consecutive edge directions,
\begin{align*}
 b_{k+1}-b_k
 &=(\gamma_{k+1}-\gamma_k)^\top u_k
   +(\gamma_{k+1}-\gamma_0)^\top(u_{k+1}-u_k)\\
 &\ge-2L\|u_{k+1}-u_k\|_2
 \ge-CL\alpha_n.
\end{align*}
Thus $b_k\ge-CLk\alpha_n$.
We telescope the forward inequalities to obtain
\begin{align*}
 r_{ji}(\theta,G)
 &=\sum_{k=0}^{N-1}
     \left[\vartheta_{k+1}-\vartheta_k
                 -\gamma_0^\top(x_{k+1}-x_k)\right]\\
 &\ge\sum_{k=0}^{N-1}s_kb_k
 \ge-CL\alpha_n\sum_{k=0}^{N-1}ks_k\\
 &\ge-CL\alpha_n N\sum_{k=0}^{N-1}s_k
 \ge-CL\frac{\alpha_n}{h_n}R^2.
\end{align*}
Substituting \eqref{eq:h-alpha} gives
$\alpha_n/h_n=(C_\alpha/c_h)v_{n,K}$, proving the claim.
\end{proof}

\subsection{Quadratic restoration of full feasibility}

Having analyzed the residual failure of the KNN solution, we next turn a KNN-feasible pair into a full-feasible comparison
point. The idea is that an appropriately small quadratic has supporting residuals proportional to
squared distance, matching the form of the possible negative
residuals in \eqref{eq:all-residual-bound}.
A single convex-combination weight can therefore properly offset all of them.

We choose $x_\ast\in\Omega$ and define
$\gamma_\ast=\min\{B/D_\Omega^2,L/(2D_\Omega)\}$ and
$q(x)=\gamma_\ast\|x-x_\ast\|_2^2/2$.
Its sampled values and slopes are
$\bar\theta_i=q(X_i)$ and $\bar g_i=\gamma_\ast(X_i-x_\ast)$.
We write $\bar z=(\bar\theta,\bar G)$ and
$c_\ast=\gamma_\ast/2>0$.

\begin{lemma}
\label{lem:quadratic-restoration}
The pair $\bar z$ is full-feasible and satisfies
\begin{equation}\label{eq:strict-comparator-residual}
 r_{ji}(\bar\theta,\bar G)
 =c_\ast\|X_i-X_j\|_2^2.
\end{equation}
For any $z=(\theta,G)\in\mathcal C$ satisfying
$r_{ji}(z)\ge-\kappa\|X_i-X_j\|_2^2$ for all pairs, with
$\kappa\ge0$, the pair
\begin{equation}\label{eq:quadratic-restoration}
 z^{\rm rep}=(1-t)z+t\bar z,
 \qquad t=\frac{\kappa}{\kappa+c_\ast},
\end{equation}
then belongs to $\mathcal C_{E_F}$.
\end{lemma}

\begin{proof}
The choice of $\gamma_\ast$ gives
$0\le\bar\theta_i\le\gamma_\ast D_\Omega^2/2\le B/2$ and
$\|\bar g_i\|_2\le\gamma_\ast D_\Omega\le L/2$.
For $a=X_j-x_\ast$ and $\Delta=X_i-X_j$, expansion gives
\begin{align*}
 r_{ji}(\bar\theta,\bar G)
 &=\frac{\gamma_\ast}{2}
       \bigl(\|a+\Delta\|_2^2-\|a\|_2^2\bigr)
       -\gamma_\ast a^\top\Delta\\
 &=\frac{\gamma_\ast}{2}\|\Delta\|_2^2.
\end{align*}
This proves \eqref{eq:strict-comparator-residual} and feasibility
of the comparison pair.

The common bound set $\mathcal C$ is convex, so $z^{\rm rep}$
satisfies the value and slope bounds. Each residual is linear in
$(\theta,G)$, and hence
\[
 r_{ji}(z^{\rm rep})
 =(1-t)r_{ji}(z)+t r_{ji}(\bar z)
 \ge[-(1-t)\kappa+tc_\ast]\|X_i-X_j\|_2^2=0.
\]
Therefore $z^{\rm rep}$ is feasible for the full problem.
\end{proof}

We apply this lemma with
$\kappa=\kappa_{n,K}$ to the KNN optimizer.
We use the restored
pair as a feasible direction in the full problem's optimality
condition.

\subsection{First-order optimality and fitted-value stability}

For notational simplicity, we write $z_F=(\theta_F,G_F)$, $z_K=(\theta_K,G_K)$, $\bar{z} = (\bar{\theta}, \bar{g})$ and
$H_n=\operatorname{diag}(I_n,\mu_nI_{nd})$ after stacking the slopes.

\begin{lemma}
\label{lem:general-optimizer-comparison}
On $\mathcal G_{n,K}$,
\begin{equation}\label{eq:general-H-stability}
 \|\theta_K-\theta_F\|_2^2
 +\mu_n\|G_K-G_F\|_F^2
 \le\frac{\kappa_{n,K}}{c_\ast}S_n,
\end{equation}
where $S_n=\langle\nabla\Phi_n(z_F),\bar z-z_F\rangle\ge0$.
\end{lemma}

\begin{proof}
We put $\Delta=z_K-z_F$.
Because $z_F\in\mathcal C_{E_K}$, KNN optimality gives
$\langle\nabla\Phi_n(z_K),z_F-z_K\rangle\ge0$.
Using $\nabla\Phi_n(z_K)=\nabla\Phi_n(z_F)+H_n\Delta$, we obtain
\begin{equation}\label{eq:general-vi-one}
 \Delta^\top H_n\Delta
 \le-\langle\nabla\Phi_n(z_F),\Delta\rangle.
\end{equation}
The quadratic pair is feasible, so full optimality implies
$S_n\ge0$.
The restored pair $z_K^{\rm rep}=(1-t)z_K+t\bar z$ is also
full-feasible. Thus
\begin{align*}
 0
 &\le\langle\nabla\Phi_n(z_F),z_K^{\rm rep}-z_F\rangle\\
 &=(1-t)\langle\nabla\Phi_n(z_F),\Delta\rangle+tS_n.
\end{align*}
Since $t/(1-t)=\kappa_{n,K}/c_\ast$,
\[
 -\langle\nabla\Phi_n(z_F),\Delta\rangle
 \le\frac{\kappa_{n,K}}{c_\ast}S_n.
\]
Combining this with \eqref{eq:general-vi-one} proves our result.
\end{proof}

The remaining comparison quantity $S_n$ depends on the full
optimizer, whose empirical error has already been bounded.
We expand it around the true fitted vector to use that bound.

\begin{lemma}[Expected comparison cost]
\label{lem:comparison-cost}
For all sufficiently large $n$,
\begin{equation}\label{eq:S-bound}
 \E_{f_0}S_n
 \le Cn\left(
 r_{n,d}+\mu_nL^2+\frac{\sigma^2}{n}
 \right)^{1/2}.
\end{equation}
Consequently, on the event $\mathcal G_{n,K}$,
\begin{equation}\label{eq:theta-KF-bound}
 \E_{f_0}\left[
 \|\theta_K-\theta_F\|_n^2
 \mathbf1_{\mathcal G_{n,K}}
 \right]
 \le Cv_{n,K}\left(
 r_{n,d}+\mu_nL^2+\frac{\sigma^2}{n}
 \right)^{1/2}.
\end{equation}
\end{lemma}

\begin{proof}
Recall that
\[
 S_n
 =
 \left\langle
 \nabla\Phi_n(\theta_F,G_F),
 (\bar\theta,\bar G)-(\theta_F,G_F)
 \right\rangle.
\]
Since $Y=\theta^0+\xi$, we can expand this quantity as
\begin{align*}
 S_n
 &= (\theta_F-\theta^0-\xi)^\top
    (\bar\theta-\theta_F)
    +\mu_n\langle G_F,\bar G-G_F\rangle\\
 &= (\theta_F-\theta^0)^\top(\bar\theta-\theta^0)
    -\|\theta_F-\theta^0\|_2^2\\
 &\quad
    -\xi^\top(\bar\theta-\theta^0)
    +\xi^\top(\theta_F-\theta^0)
    +\mu_n\langle G_F,\bar G-G_F\rangle.
\end{align*}

We first control the terms involving the fitted values.
Since $\bar\theta$ and $\theta^0$ are functions of the
design $X$, and the noise is conditionally centered,
\[
 \E_{f_0}
 \left[\xi^\top(\bar\theta-\theta^0)\right]=0.
\]
Moreover, the bounds $|\bar\theta_i|\le B$ and
$|\theta_i^0|\le B$ give us
$\|\bar\theta-\theta^0\|_2\le2B\sqrt n.$
We know from Proposition A.14 that $\E_{f_0}\|\theta_F-\theta^0\|_2^2
 \le Cn\left(
 r_{n,d}+\mu_nL^2+\frac{\sigma^2}{n}
 \right).$
Therefore, Cauchy--Schwarz yields
\begin{align*}
 &\E_{f_0}\left[
 (\theta_F-\theta^0)^\top
 (\bar\theta-\theta^0)\right]\\
 &\quad\le
 \left(
 \E_{f_0}\|\theta_F-\theta^0\|_2^2
 \right)^{1/2}
 \left(
 \E_{f_0}\|\bar\theta-\theta^0\|_2^2
 \right)^{1/2}\\
 &\quad\le Cn\left(
 r_{n,d}+\mu_nL^2+\frac{\sigma^2}{n}
 \right)^{1/2}.
\end{align*}

Similarly, Lemma \ref{lem:subgaussian-elementary} gave us
$\E\|\xi\|_2^2\le n\sigma^2$. Another application of Cauchy Schwartz then says that
\begin{align*}
 &\E_{f_0}\left[
 \xi^\top(\theta_F-\theta^0)\right]\le
 \left(\E\|\xi\|_2^2\right)^{1/2}
 \left(
 \E_{f_0}\|\theta_F-\theta^0\|_2^2
 \right)^{1/2}\le C\sigma n\left(
 r_{n,d}+\mu_nL^2+\frac{\sigma^2}{n}
 \right)^{1/2}.
\end{align*}

For the regularization term, the slope bounds give
\[
 \mu_n\langle G_F,\bar G-G_F\rangle
 \le 2\mu_n nL^2.
\]
We can also discard the nonpositive term
$-\|\theta_F-\theta^0\|_2^2$. Combining these
estimates gives
\[
 \E_{f_0}S_n
 \le Cn(1+\sigma)
 \left(
 r_{n,d}+\mu_nL^2+\frac{\sigma^2}{n}
 \right)^{1/2}
 +2\mu_n nL^2.
\]
By Assumption~\ref{ass:regularization},
$\mu_nL^2\le C_\mu r_{n,d}$. Since $r_{n,d}\to0$,
for all sufficiently large $n$,
\[
 \mu_nL^2
 \le C\left(
 r_{n,d}+\mu_nL^2+\frac{\sigma^2}{n}
 \right)^{1/2}.
\]
Absorbing fixed constants into $C$ establishes
\eqref{eq:S-bound}.

Finally, Eq. \eqref{eq:general-H-stability} gives that
on $\mathcal G_{n,K}$,
\[
 \|\theta_K-\theta_F\|_2^2
 \le \frac{\kappa_{n,K}}{c_\ast}S_n.
\]
Since $\kappa_{n,K}\le Cv_{n,K}$, we can divide
by $n$, multiply by $\mathbf1_{\mathcal G_{n,K}}$,
and take expectations to obtain
\[
 \E_{f_0}\left[
 \|\theta_K-\theta_F\|_n^2
 \mathbf1_{\mathcal G_{n,K}}
 \right]
 \le \frac{Cv_{n,K}}{n}
 \E_{f_0}\left[
 S_n\mathbf1_{\mathcal G_{n,K}}
 \right].
\]
Because $S_n\ge0$ by first-order optimality of the
full estimator, we have
\[
 \E_{f_0}\left[
 S_n\mathbf1_{\mathcal G_{n,K}}
 \right]
 \le \E_{f_0}S_n.
\]
Substituting \eqref{eq:S-bound} proves
\eqref{eq:theta-KF-bound}.
\end{proof}

\subsection{Population comparison and the neighborhood choice}

The fitted-value comparison is not yet a prediction comparison:
other affine pieces can place $\widehat f_K(X_i)$ above
$\theta_{K,i}$. Thankfully, the omitted-residual estimate also controls this
interpolation defect, allowing us to complete the proof.

\begin{proof}[Proof of Theorem~\ref{thm:general-detailed}]
We define the interpolation defect $M_i=\widehat f_K(X_i)-\theta_{K,i} = \max_{j}\{-r_{ji}(\theta_K, G_K)\}$.
By \eqref{eq:knn-interpolation-defect} and
Lemma~\ref{lem:omitted-residual-control}, on $\mathcal G_{n,K}$,
\[
 0\le M_i\le\kappa_{n,K}D_\Omega^2,
 \qquad \frac1n\sum_iM_i^2\le Cv_{n,K}^2.
\]
For the full convex regression problem, we have that $f_{F}(X_i) = \theta_{F, i}$ implying that
$\widehat f_K(X_i)-\widehat f_F(X_i)
=(\theta_{K,i}-\theta_{F,i})+M_i$.
Using $(a+b)^2\le2a^2+2b^2$ and
\eqref{eq:theta-KF-bound}, we obtain
\begin{equation}\label{eq:sample-prediction-diff}
 \E_{f_0}\left[
   \frac1n\sum_i
   (\widehat f_K(X_i)-\widehat f_F(X_i))^2
          \mathbf1_{\mathcal G_{n,K}}\right]
 \le C(v_{n,K}
\left(r_{n,d}+\mu_nL^2+\frac{\sigma^2}{n}\right)^{1/2}+v_{n,K}^2).
\end{equation}
Both predictors are $L$-Lipschitz. On $\mathcal G_{n,K}$,
Lemma~\ref{lem:population_to_sample} and
the bounds for $p_{max}$ and $\rho_n$ therefore give
\begin{align*}
 &\E_{f_0}\left[
   \ell_{\mathbb P}^2(\widehat f_K,\widehat f_F)
            \mathbf1_{\mathcal G_{n,K}}\right]\\
 &\qquad\le C\left[
   \ell_n(v_{n,K}\left(r_{n,d}+\mu_nL^2+\frac{\sigma^2}{n}\right)^{1/2} +v_{n,K}^2)
       +(\ell_n/n)^{2/d}\right].
\end{align*}
On the complementary event, the squared discrepancy is at most
$4B_f^2$, and the event has probability at most $\delta_n$.
Adding this contribution proves
\eqref{eq:general-detailed-bound}, for any $f_0$ in the function class.
\end{proof}

We know balance the terms in \ref{thm:general-detailed} through our choice of $K$.

\begin{corollary}
\label{cor:general-neighborhoods}
For a sufficiently large fixed constant $C$, the deterministic choice
\begin{equation}\label{eq:general-K-all-d}
 K_n=\left\lceil
 C n^{\frac{d-1}{2d-1}}
   \ell_n^{\frac d{2d-1}}
   r_{n,d}^{-\frac{d(d-1)}{2(2d-1)}}
 \right\rceil
\end{equation}
satisfies $K_n=o(n)$ and, for all sufficiently large $n$,
\begin{equation}\label{eq:general-risk-transfer}
 \sup_{f_0\in\mathcal F_{B,L}(\Omega)}
 \E_{f_0}\left[
  \ell_{\mathbb P}^2(\widehat f_K,f_0)
  \right]
 \le Cr_{n,d}\log n.
\end{equation}
Additionally, 
\begin{align}
     \sup_{f_0\in\mathcal F_{B,L}(\Omega)}
 \E_{f_0}\left[
  \ell_{\mathbb P}^2(\widehat f_K,\widehat f_F)
  \right]
 \le Cr_{n,d}\log n.
\end{align}
\end{corollary}

\begin{proof}
Assumption \ref{ass:regularization} gives us that
$\left(r_{n,d}+\mu_nL^2+\frac{\sigma^2}{n}\right)\le Cr_{n,d}$. Properly expanding our $v_{n,K}$ gives us that:
\[
 v_{n,K}
 =n^{1/d}\ell_n^{1/(d-1)}
       K^{-(2d-1)/[d(d-1)]}.
\]
Thus $v_{n,K}\le c r_{n,d}^{1/2}$ is obtained by taking a
sufficiently large constant in \eqref{eq:general-K-all-d}.
For this choice,
$$\ell_n(v_{n,K}\left(r_{n,d}+\mu_nL^2+\frac{\sigma^2}{n}\right)+v_{n,K}^2)\le Cr_{n,d}\log n$$.

We now need to establish that $(\ell_n/n)^{2/d} \leq Cr_{n, d} \log n$ Noting that $\ell_n$ is on the same order as $\log n$ and checking each dimension suffices to prove this.

Finally, we need to verify that our chosen values of $K_n$ satisfy the condition of $C_0 l_n \leq K_n \leq c_0 n$.
Since $\ell_n\asymp\log n$, substituting the expressions for $r_{n,d}$ gives us
\begin{equation}\label{eq}
K_n\asymp
\begin{cases}
n^{\frac{(d-1)(3d+4)}{(2d-1)(d+4)}}
(\log n)^{\frac d{2d-1}},
&2\le d<4,\\
n^{6/7}(\log n)^{-2/7},
&d=4,\\
n^{\frac{2(d-1)}{2d-1}}
(\log n)^{\frac d{2d-1}},
&d>4.
\end{cases}
\end{equation}
In the first case, the polynomial exponent equals $5/9$ for $d=2$ and $26/35$ for $d=3$. For $d=4$, it equals $6/7$, while for $d>4$ it is $2(d-1)/(2d-1)<1$. Thus, in every fixed dimension $d\ge2$, the neighborhood size has the form $K_n\asymp n^p(\log n)^q$ for some $0<p<1$ and fixed $q\in\mathbb R$.

It follows that
\[
\frac{K_n}{\ell_n}
\asymp n^p(\log n)^{q-1}\longrightarrow\infty,
\qquad
\frac{K_n}{n}
\asymp n^{p-1}(\log n)^q\longrightarrow0,
 \]

We then have that $\mathbb E_{f_0}\ell_{\mathbb P}^2(\widehat f_K,\widehat f_F)
\le Cr_{n,d}\log n.$ Finally, the risk decomposition with Proposition \eqref{prop:regularized-full-risk} gives
$$
\mathbb E_{f_0}\ell_{\mathbb P}^2(\widehat f_K,f_0)\le
2\mathbb E_{f_0}\ell_{\mathbb P}^2(\widehat f_K,\widehat f_F)+
2\mathbb E_{f_0}\ell_{\mathbb P}^2(\widehat f_F,f_0)\le Cr_{n,d}\log n.$$ where $C$ is appropriately increased in the last inequality
\end{proof}

\section{Strongly Convex Case}
\label{app:strong-proof}

We move onto the more specific case of strongly convex functions. We prove Theorem~\ref{thm:strong-summary} by identifying a radius
that contains every active constraint of the full problem.
We first strengthen the full estimator's population-error control
to an error bound holding everywhere on $\Omega$.
Strong convexity then makes every sufficiently distant constraint
strictly slack. Therefore, a KNN graph retaining all shorter constraints has
the same optimizer as the full problem completing our proof. 
\subsection{Uniform approximation from population error}
\label{subsec:strong-convex-bias}

We begin by showing that Lipschitzness gives us a way to bound uniform approximation from $L_2$ error.  
\begin{lemma}
\label{lem:l2-to-sup}
For an $A$-Lipschitz function $h:\Omega\to\mathbb R$,
\begin{equation}\label{eq:l2-to-sup}
 \|h\|_\infty
 \le C_\Omega\left[
    \left(\int_\Omega h^2\,d\mathbb P\right)^{1/2}
    +A^{d/(d+2)}
     \left(\int_\Omega h^2\,d\mathbb P\right)^{1/(d+2)}
 \right].
\end{equation}
\end{lemma}

\begin{proof}
If $\|h\|_\infty=0$, the conclusion is immediate. If $A=0$,
then $h$ is constant on $\Omega$, so
$\|h\|_\infty
=
\left(\int_\Omega h^2\,d\mathbb P\right)^{1/2}.
$
We therefore assume that $\|h\|_\infty>0$ and $A>0$.

Since $\Omega$ is compact and $h$ is continuous, there exists
$x_0\in\Omega$ such that $|h(x_0)|=\|h\|_\infty$. By Lipschitz
continuity, for every $x\in\Omega$,
\[
|h(x)|
\ge |h(x_0)|-|h(x)-h(x_0)|
\ge \|h\|_\infty-A\|x-x_0\|_2.
\]
In particular, whenever
$\|x-x_0\|_2\le \|h\|_\infty/(2A)$, we have
$|h(x)|\ge\|h\|_\infty/2$.

We first consider the case
$\|h\|_\infty/(2A)\le D_\Omega$.
Applying the ball-mass bound~\eqref{eq:ball-masses}, we obtain
\begin{align*}
\int_\Omega h^2\,d\mathbb P
&\ge
\int_{B(x_0,\|h\|_\infty/(2A))\cap\Omega}
h^2\,d\mathbb P\\
&\ge
\frac{\|h\|_\infty^2}{4}
\mathbb P\left(
B\left(x_0,\frac{\|h\|_\infty}{2A}\right)
\cap\Omega
\right)\\
&\ge
c_\Omega\frac{\|h\|_\infty^2}{4}
\left(\frac{\|h\|_\infty}{2A}\right)^d\\
&=
\frac{c_\Omega}{2^{d+2}}
\frac{\|h\|_\infty^{d+2}}{A^d}.
\end{align*}
Rearranging yields
$
\|h\|_\infty
\le
C_\Omega A^{d/(d+2)}
\left(\int_\Omega h^2\,d\mathbb P\right)^{1/(d+2)}.
$

We next consider the case
$\|h\|_\infty/(2A)>D_\Omega$.
Since $\|x-x_0\|_2\le D_\Omega$ for every
$x\in\Omega$, the Lipschitz inequality gives
\[
|h(x)|
\ge\|h\|_\infty-AD_\Omega
>\frac{\|h\|_\infty}{2}
\qquad\text{for all }x\in\Omega.
\]
Consequently,
$
\int_\Omega h^2\,d\mathbb P
\ge\frac{\|h\|_\infty^2}{4},
\implies 
\|h\|_\infty
\le2\left(\int_\Omega h^2\,d\mathbb P\right)^{1/2}.
$

Combining the two cases and increasing $C_\Omega$ if necessary,
we conclude that
\[
\|h\|_\infty
\le C_\Omega\left[
\left(\int_\Omega h^2\,d\mathbb P\right)^{1/2}
+
A^{d/(d+2)}
\left(\int_\Omega h^2\,d\mathbb P\right)^{1/(d+2)}
\right].
\]
\end{proof}

We apply this deterministic lemma to the error
$\widehat f_F-f_0$, which is $2L$-Lipschitz.
The population confidence bound from Appendix~A then provides a
single event on which the full predictor is accurate at every
point, including the midpoints used in the screening argument.

\begin{lemma}[Uniform approximation of the full predictor]
\label{lem:uniform-full-screening}
For a sufficiently large fixed constant $C$ and all sufficiently
large $n$,
\begin{equation}\label{eq:uniform-full-hp}
\sup_{f_0\in\mathcal F_{B,L}(\Omega)}
\PP_{f_0}\left\{
\|\widehat f_F-f_0\|_\infty>\epsilon_n
\right\}
\le\delta_n,
\end{equation}
where
$
\epsilon_n
=
C\left(r_{n,d}\log n\right)^{1/(d+2)},
$ and $\delta_n=r_{n,d}^2.
$
\end{lemma}

\begin{proof}
We apply the population confidence bound
\eqref{eq:full-population-confidence} probability of 
$\delta_n$. Since $\delta_n=r_{n,d}^2$ and
$\log(n/\delta_n)\le C\log n$ for all sufficiently large $n$,
we get
\[
\sup_{f_0\in\mathcal F_{B,L}(\Omega)}
\PP_{f_0}\left\{
\ell_{\mathbb P}^2(\widehat f_F,f_0)
>
C r_{n,d}\log n
\right\}
\le\delta_n.
\]

Both $\widehat f_F$ and $f_0$ are $L$-Lipschitz, so their
difference is $2L$-Lipschitz. So, on the event
where $
\ell_{\mathbb P}^2(\widehat f_F,f_0)
\le C r_{n,d}\log n,
$, Lemma~\ref{lem:l2-to-sup} gives
\begin{align*}
\|\widehat f_F-f_0\|_\infty
&\le C_\Omega\left[
\left(Cr_{n,d}\log n\right)^{1/2}
+(2L)^{d/(d+2)}
\left(Cr_{n,d}\log n\right)^{1/(d+2)}
\right]\\
&\le C\left(r_{n,d}\log n\right)^{1/(d+2)}.
\end{align*}
The last inequality follows because
$r_{n,d}\log n\to0$ and
$1/(d+2)<1/2$, so
\[
\left(r_{n,d}\log n\right)^{1/2}
\le
\left(r_{n,d}\log n\right)^{1/(d+2)}
\]
for all sufficiently large $n$.

Choosing the constant in $\epsilon_n$ sufficiently large
establishes \eqref{eq:uniform-full-hp}, uniformly over
$f_0\in\mathcal F_{B,L}(\Omega)$.
\end{proof}

\subsection{Screening active supporting constraints}

We now assume $f_0\in\mathcal F^{\rm sc}_{B,L}(\Omega;m_0)$
for a fixed $m_0>0$.
A full constraint is active when its residual is zero: the affine
piece based at $X_j$ then agrees with the full predictor at both
$X_j$ and $X_i$.
Strong convexity of the truth creates a definite midpoint gap
between distant observations. If the full predictor uniformly
approximates the truth, that gap prevents a distant constraint
from remaining active.

\begin{lemma}
\label{lem:strong-convexity-screens}
On the event $\|\widehat f_F-f_0\|_\infty\le\epsilon_n$,
\begin{equation}\label{eq:screening-inequality}
 r_{ji}(\theta_F,G_F)
 \ge\frac{m_0}{4}\|X_i-X_j\|_2^2-4\epsilon_n,
 \qquad i\ne j.
\end{equation}
Consequently, every active full constraint has length at most
\begin{equation}\label{eq:screening-radius}
 R_n=4\sqrt{\epsilon_n/m_0}.
\end{equation}
\end{lemma}

\begin{proof}
We set $h=\widehat f_F$, $x=X_j$, $y=X_i$, and $z=(x+y)/2$.
Since we are looking at the full solution of the convex problem, the $j$th affine piece is a supporting
plane of $h$ at $x$. Evaluating it at the midpoint gives
$h(z)\ge h(x)+g_{F,j}^\top(y-x)/2$.
Therefore,
\begin{align*}
 r_{ji}(\theta_F,G_F)
 &=h(y)-h(x)-g_{F,j}^\top(y-x)\\
 &\ge h(y)+h(x)-2h(z)\\
 &\ge f_0(y)+f_0(x)-2f_0(z)-4\epsilon_n.
\end{align*}
Using the definition of strong convexity at the midpoints ($t=1/2$) in
gives
\[
 f_0(y)+f_0(x)-2f_0((x+y)/2)
 \ge\frac{m_0}{4}\|y-x\|_2^2.
\]
This proves \eqref{eq:screening-inequality}.
If $\|y-x\|_2>R_n$, its right-hand side is positive, so the
constraint is strictly slack. Every active constraint must
therefore have length at most $R_n$.
\end{proof}

Once all potentially active constraints have been retained, a
separate deterministic fact completes the optimization argument.
Removing only inequalities that are strictly slack at the full
optimizer leaves that optimizer unchanged.

\subsection{Exact recovery and population-risk transfer}

The radius $R_n$ is deterministic and tends to zero.
The local-count lemma bounds how many observations can lie within
this radius of any source. So our strategy is to choose $K_n$ to retain them all.
On the resulting event, the preceding lemmas give exact agreement;
bounded losses then control the contribution from the exceptional
datasets.

\begin{theorem}[Strongly convex sparsification preserves the full-estimator rate]
\label{thm:main}
Under the setup of Section~\ref{sec:framework} and
Assumption~\ref{ass:regularization}, we consider a nonempty class
$\mathcal F^{\rm sc}_{B,L}(\Omega;m_0)$ with fixed $m_0>0$.
For all sufficiently large $n$, any deterministic integer
$1\le K_n\le n-1$ satisfying
\begin{equation}\label{eq:main-K}
 K_n\ge C\left[
 n\,m_0^{-d/2}
 (r_{n,d}\log n)^{\frac{d}{2(d+2)}}
 +\log n
 \right]
\end{equation}
for a sufficiently large fixed constant $C$ satisfies
\begin{equation}\label{eq:screening-agreement-probability}
 \sup_{f_0\in\mathcal F^{\rm sc}_{B,L}(\Omega;m_0)}
 \PP_{f_0}\{(\theta_K,G_K)\ne(\theta_F,G_F)\}
 \le Cr_{n,d}^2
\end{equation}
and
\begin{equation}\label{eq:bias-negligible}
 \sup_{f_0\in\mathcal F^{\rm sc}_{B,L}(\Omega;m_0)}
 \E_{f_0}\ell_{\mathbb P}^2(\widehat f_K,\widehat f_F)
 \le Cr_{n,d}^2.
\end{equation}
We also have \begin{equation}\label{eq:exact-risk-transfer}
 |\mathcal R_{K,n}-\mathcal R_{F,n}|
 \le Cr_{n,d}^2,
\end{equation}
and consequently
\begin{equation}\label{eq:knn-rate-preservation}
 \mathcal R_{K,n}\le Cr_{n,d}\log n.
\end{equation}
There are choices satisfying \eqref{eq:main-K} with $K_n=o(n)$.
\end{theorem}

\begin{proof}
Keeping the same definitions for $\epsilon_n$ and $R_n$
from  Lemma~\ref{lem:uniform-full-screening} and Lemma \ref{lem:strong-convexity-screens}, we consider the event $\mathcal U_n
 =
 \{\|\widehat f_F-f_0\|_\infty\le\epsilon_n\}.$
Lemma~\ref{lem:uniform-full-screening} gives
$
 \sup_{f_0\in\mathcal F^{\rm sc}_{B,L}(\Omega;m_0)}
 \PP_{f_0}(\mathcal U_n^c)
 \le r_{n,d}^2.$
On $\mathcal U_n$, Lemma~\ref{lem:strong-convexity-screens}
shows that every active full supporting-hyperplane constraint
has length at most $R_n$.

Since $r_{n,d}\log n\to0$, we have
$\epsilon_n\to0$ and hence $R_n\to0$. Thus
$R_n\le D_\Omega$ for all sufficiently large $n$.
We apply Lemma~\ref{lem:local-counts} at the deterministic
radius $R_n$, with failure probability $r_{n,d}^2$.
On the resulting event $\mathcal G_n$,
\[
 \max_{j\le n}
 \#\{i\ne j:\|X_i-X_j\|_2\le R_n\}
 \le C_N\left[
 nR_n^d+\log\left(\frac{n}{r_{n,d}^2}\right)
 \right].
\]
Since $\log(n/r_{n,d}^2)\le C\log n$ for all sufficiently
large $n$, and
\begin{align*}
 nR_n^d
 &=4^d n m_0^{-d/2}\epsilon_n^{d/2}\\
 &\le Cn m_0^{-d/2}
 (r_{n,d}\log n)^{\frac{d}{2(d+2)}},
\end{align*}
condition~\eqref{eq:main-K} ensures that every observation
within distance $R_n$ of any $X_j$ belongs to its
$K_n$ nearest neighbors.

Consequently, on $\mathcal U_n\cap\mathcal G_n$,
the KNN graph retains every active constraint of the full
problem. Every omitted full constraint is therefore strictly
slack at $(\theta_F,G_F)$ and removing these constraints
does not change the optimizer, giving us
\begin{equation}\label{eq:exact-recovery-on-good-event}
 (\theta_K,G_K)=(\theta_F,G_F)
 \qquad\text{on }\mathcal U_n\cap\mathcal G_n.
\end{equation}
The union bound gives
\[
 \sup_{f_0\in\mathcal F^{\rm sc}_{B,L}(\Omega;m_0)}
 \PP_{f_0}\bigl((\mathcal U_n\cap\mathcal G_n)^c\bigr)
 \le 2r_{n,d}^2,
\]
which proves \eqref{eq:screening-agreement-probability}. On the $\mathcal U_n\cap\mathcal G_n$, the predictors coincide everywhere.
On its complement, both predictors are uniformly bounded by
$B_f=B+LD_\Omega$, so
$
 \ell_{\mathbb P}^2(\widehat f_K,\widehat f_F)
 \le4B_f^2.
$
Therefore,
\begin{align*}
 \E_{f_0}\ell_{\mathbb P}^2(\widehat f_K,\widehat f_F)
 &\le
 4B_f^2
 \PP_{f_0}\bigl((\mathcal U_n\cap\mathcal G_n)^c\bigr)\\
 &\le8B_f^2r_{n,d}^2.
\end{align*}
Taking the supremum over the truth class establishes
\eqref{eq:bias-negligible}.

We next compare the population risks against $f_0$.
On the agreement event, the two predictors and hence their
squared population losses coincide. On the complementary event,
both losses are bounded by $(B_f+B)^2$. Thus,
\begin{align*}
 &\left|
 \E_{f_0}\ell_{\mathbb P}^2(\widehat f_K,f_0)
 -
 \E_{f_0}\ell_{\mathbb P}^2(\widehat f_F,f_0)
 \right|\\
 &\qquad\le
 (B_f+B)^2
 \PP_{f_0}\bigl((\mathcal U_n\cap\mathcal G_n)^c\bigr)\\
 &\qquad\le
 2(B_f+B)^2r_{n,d}^2.
\end{align*}
Taking the supremum over the truth class and using
\[
 \left|\sup_f a_f-\sup_f b_f\right|
 \le\sup_f|a_f-b_f|
\]
proves \eqref{eq:exact-risk-transfer}.
Combining this bound with
Proposition~\ref{prop:regularized-full-risk} gives
\[
 \mathcal R_{K,n}
 \le\mathcal R_{F,n}+Cr_{n,d}^2
 \le Cr_{n,d}\log n,
\]
which establishes \eqref{eq:knn-rate-preservation}. Finally, we verify that the neighborhood condition admits
sublinear choices. Since $r_{n,d}\log n\to0$,
\[
 m_0^{-d/2}
 (r_{n,d}\log n)^{\frac{d}{2(d+2)}}
 +\frac{\log n}{n}
 \longrightarrow0.
\]
Thus, taking $K_n$ to be the ceiling of a sufficiently
large constant multiple of the right-hand side of
\eqref{eq:main-K} gives $K_n/n\to0$.
\end{proof}

Plugging in the values of $r_{n, d}$ then gives us that
\begin{corollary}
\label{cor:explicit-rates}
For fixed $m_0>0$, sufficiently large constant multiples of the
following scales, rounded upward and capped at $n-1$, satisfy
Theorem~\ref{thm:main} for all sufficiently large $n$:
\begin{equation}\label{eq:K-all-dimensions}
 K_n = \begin{cases}
 n^{1-\frac{2d}{(d+4)(d+2)}}
       (\log n)^{\frac{d}{2(d+2)}},&2\le d<4,\\[0.4em]
 n^{5/6}(\log n)^{2/3},&d=4,\\[0.4em]
 n^{\frac{d+1}{d+2}}
       (\log n)^{\frac{d}{2(d+2)}},&d>4.
 \end{cases}
\end{equation}
For these choices,
\begin{equation}\label{eq:explicit-rate-summary}
 \mathcal R_{K,n}\le Cr_{n,d}\log n,
 \qquad
 |\mathcal R_{K,n}-\mathcal R_{F,n}|\le Cr_{n,d}^2,
 \qquad nK_n=o(n^2).
\end{equation}
\end{corollary}

\begin{proof}
The fixed factor $m_0^{-d/2}$ is included in that constant.

For $2\le d<4$, the leading term becomes
\[
 n\bigl(n^{-4/(d+4)}\log n\bigr)^{d/[2(d+2)]}
 =n^{1-\frac{2d}{(d+4)(d+2)}}
       (\log n)^{\frac d{2(d+2)}}.
\]
For $d=4$, it is
$n\{n^{-1/2}(\log n)^2\}^{1/3}
=n^{5/6}(\log n)^{2/3}$.
For $d>4$, it is
\[
 n\bigl(n^{-2/d}\log n\bigr)^{d/[2(d+2)]}
 =n^{\frac{d+1}{d+2}}(\log n)^{\frac d{2(d+2)}}.
\]
Each polynomial exponent is strictly between zero and one, so the
additive $\log n$ is lower order and each choice has $K_n=o(n)$.
\end{proof}
\section{Optimization Procedures}
\label{app:optimization}
\label{sec:optimization}

We gives the relevant optimization formulation, our used numerical
method, and the stopping rules used in our experiments. The same method
handles the full and KNN estimators. We first define the problem and augmented Lagrangian,
then explain the inner Newton solves, acceptance tests, and continuation
between problems. The final algorithm summarizes these steps. 

\subsection{Regression problem and notation}
\label{subsec:operator}
Given observations $(X_i,Y_i)\in\mathbb R^d\times\mathbb R$, $i=1,\ldots,n$,
write $Y=(Y_1,\ldots,Y_n)^\top$ and let $W$ be a set of directed pairs
$(j,i)$ with $j\ne i$. The source $j$ specifies a supporting plane and the
target $i$ specifies where it is constrained. Collect the unknown fitted
values in $\theta=(\theta_1,\ldots,\theta_n)^\top$ and slopes in
$G=(g_1,\ldots,g_n)$, with $g_i\in\mathbb R^d$. We solve
\begin{equation}\label{eq:operator-sign}
\begin{aligned}
 \min_{\theta,G}\quad
 \Phi_n(\theta,G)&=\frac12\sum_{i=1}^n(\theta_i-Y_i)^2
                 +\frac{\mu_n}{2}\sum_{i=1}^n\|g_i\|_2^2,\\
 \text{subject to}\quad
 \theta_j-\theta_i+g_j^\top(X_i-X_j)&\leq0,\qquad (j,i)\in W,\\
 |\theta_i|&\leq B,\quad \|g_i\|_2\leq L,\qquad i=1,\ldots,n.
\end{aligned}
\end{equation}
For notational simplicity, we stack the column vectors as $w=(\theta^\top,g_1^\top,\ldots,g_n^\top)^\top\in\mathbb R^{p}$, with
$p=n(d+1)$, and define
\[
 t=(Y,0),\qquad H=\operatorname{diag}(I_n,\mu_n I_{nd}),\qquad
 (C_Ww)_{ji}=\theta_j-\theta_i+g_j^\top(X_i-X_j).
\]
Here $I_q$ is the $q\times q$ identity, $t$ appends $nd$ zeros to $Y$,
and $C_W$ is a linear map from $\mathbb R^p$ to $\mathbb R^{|W|}$. Then, we can write our objection function
$\Phi_n(w)=\tfrac12(w-t)^\top H(w-t)$
$C_Ww$ are residual inequalities. Let
$\mathcal C=\{w:|\theta_i|\leq B,\ \|g_i\|_2\leq L\}$.
Its Euclidean projection $\Pi_{\mathcal C}$ clips each fitted value to
$[-B,B]$ and maps each slope $g_i$ to
$g_i\min\{1,L/\|g_i\|_2\}$, leaving zero slopes unchanged.

\subsection{Augmented Lagrangian updates}
\label{subsec:alm-implementation}
We introduce an auxiliary edge vector $z\in\mathbb R^{|W|}$ and a copy
$v\in\mathbb R^p$ of the fitted values and slopes. Write
$\mathbb R^{|W|}_-=\{z:z_e\leq0\text{ for every }e\}$, and let
$\delta_S(u)$ equal zero when $u\in S$ and $+\infty$ otherwise.
The convex regression problem is equivalently
\begin{equation}\label{eq:alm-split-problem}
 \min_{w,z,v}\ \Phi_n(w)+\delta_{\mathbb R^{|W|}_-}(z)+\delta_{\mathcal C}(v)
 \quad\text{subject to}\quad C_Ww-z=0,\quad w-v=0.
\end{equation}
The two auxiliary variables separate the simple sign and bound constraints
from the linear equalities linking them to $w$.

For equality multipliers $\lambda\in\mathbb R^{|W|}$ and
$\nu\in\mathbb R^p$, and penalty $\sigma>0$, define
\begin{equation}\label{eq:augmented-lagrangian}
\begin{aligned}
 \mathcal L_\sigma(w,z,v;\lambda,\nu)
 ={}&\Phi_n(w)+\delta_{\mathbb R^{|W|}_-}(z)+\delta_{\mathcal C}(v)\\
 &+\lambda^\top(C_Ww-z)+\nu^\top(w-v)\\
 &+\frac{\sigma}{2}\|C_Ww-z\|_2^2
  +\frac{\sigma}{2}\|w-v\|_2^2.
\end{aligned}
\end{equation}

At outer iteration $k$, we hold $\lambda^k,\nu^k,\sigma_k$ fixed and minimize
\[
 \mathcal L_{\sigma_k}(w,z,v;\lambda^k,\nu^k)
       +\frac{\eta}{2\sigma_k}\|w-w^k\|_2^2,
\]
for a set $\eta$. For any fixed $w$, the minimizing $z$ and $v$ are available explicitly.
For example, completing the square gives
\[
 (\lambda^k)^\top(C_Ww-z)+\frac{\sigma_k}{2}\|C_Ww-z\|_2^2
 =\frac{\sigma_k}{2}\|C_Ww+\lambda^k/\sigma_k-z\|_2^2
   -\frac{\|\lambda^k\|_2^2}{2\sigma_k}.
\]
Thus minimizing over $z\leq0$ is a projection onto the negative orthant;
the same calculation for $v\in\mathcal C$ gives
\begin{equation}\label{eq:alm-auxiliary-projections}
\begin{aligned}
 z_k(w)&=\min\{C_Ww+\lambda^k/\sigma_k,0\},\\
 v_k(w)&=\Pi_{\mathcal C}(w+\nu^k/\sigma_k),
\end{aligned}
\end{equation}
where the minimum is componentwise. By denoting $\operatorname{dist}(u,\mathcal C)=\|u-\Pi_{\mathcal C}u\|_2$.
Substituting these exact minimizers
leaves the following subproblem in $w$ alone:
\begin{equation}\label{eq:alm-subproblem}
\begin{aligned}
 \psi_k(w)={}&\Phi_n(w)
  +\frac{\|[\lambda^k+\sigma_k C_Ww]_+\|_2^2}{2\sigma_k}\\
 &+\frac{\sigma_k}{2}\operatorname{dist}^2
      (w+\nu^k/\sigma_k,\mathcal C)
  +\frac{\eta}{2\sigma_k}\|w-w^k\|_2^2.
\end{aligned}
\end{equation}
The omitted constant is
$-(\|\lambda^k\|_2^2+\|\nu^k\|_2^2)/(2\sigma_k)$, which has no effect
on the minimizing $w$. The proximal term stabilizes the iteration around
$w^k$; it does not change the regression objective. Semismooth Newton is
used to minimize this reduced function $\psi_k$.

\paragraph{Updating the multipliers.}
Minimizing \eqref{eq:alm-subproblem} gives us a new update $w^{k+1}$.
We then update $v^{k+1}, \lambda^{k+1}$, and $\nu^{k+1}$ with:
\begin{align}\label{eq:alm-updates}
 v^{k+1}&=\Pi_{\mathcal C}(w^{k+1}+\nu^k/\sigma_k),\notag\\
 \lambda^{k+1}&=[\lambda^k+\sigma_k C_Ww^{k+1}]_+,\\
 \nu^{k+1}&=\nu^k+\sigma_k(w^{k+1}-v^{k+1}).\notag
\end{align}
The projection formula makes $\lambda^{k+1}$ nonnegative, even though
equality multipliers are unrestricted in the split formulation.
The vector $v^{k+1}$ supplies the current bound-constrained model.
To adjust the penalty, define
\[
 r_{k+1}^2=
 \frac{\|\lambda^{k+1}-\lambda^k\|_2^2}{\sigma_k^2}
 +\|w^{k+1}-v^{k+1}\|_2^2,\qquad r_0=\infty.
\]
If $r_{k+1}>0.7r_k$, we set
$\sigma_{k+1}=\min\{2\sigma_k,10^5\}$; otherwise we retain $\sigma_k$.
This residual controls penalty adjustment. Final acceptance uses the
objective or risk prediction tests below.
\subsection{Objective and prediction stopping rules}

We define the stopping of our algorithm with two procedures:

\label{subsec:certificates}
\paragraph{Objective convergence.}
Let $\Phi_n^*(W)$ denote the minimum objective in
\eqref{eq:operator-sign}. For any $\lambda\geq0$, define the bounded
Lagrangian value
\[
 q_W(\lambda)=\min_{u\in\mathcal C}
                  \{\Phi_n(u)+\lambda^\top C_Wu\}.
\]
At any feasible $u$, the multiplier term is nonpositive, so
$q_W(\lambda)\leq\Phi_n^*(W)$. Completing the square gives its minimizer
and value explicitly:
\begin{equation}\label{eq:dual-value}
\begin{aligned}
 w(\lambda)&=\Pi_{\mathcal C}(t-H^{-1}C_W^\top\lambda),\\
 q_W(\lambda)&=\Phi_n(w(\lambda))+\lambda^\top C_Ww(\lambda).
\end{aligned}
\end{equation}
The Euclidean projection in the first equality is valid because $H$ is
scalar on each value and slope block. We retain the largest encountered value
$D=q_W(\lambda)$ and the smallest objective $U=\Phi_n(w^{\rm feas})$ of a
model satisfying every target constraint. Thus $D\leq\Phi_n^*(W)\leq U$.
Initially $D=q_W(0)$ and $U=\Phi_n(0)$; subsequently we keep the best bound
of each kind, together with its generating multiplier or model.
After each completed outer update, we define our algorithm to have objective convergence if
\begin{equation}\label{eq:experiment-gap-acceptance}
 (U-D)/n\leq10^{-4}.
\end{equation}

\paragraph{Risk convergence.}\label{subsec:risk-stopping}
A second stopping route addresses our prediction comparison directly. We used
$m=4096$ independent monitoring points $Z_j$ drawn from the design
distribution and their noiseless truth values $f_0(Z_j)$. This sample is
fixed throughout a fit and is independent of training and final testing.
After each completed outer update of our augmented Lagrangian method,
we evaluate the candidate predictor $f_\ell$ and its monitoring risk
\[
 \widehat R_\ell=\frac1m\sum_{j=1}^m(f_\ell(Z_j)-f_0(Z_j))^2.
\]

Let $\mathcal I_\ell$ index the latest five consecutive valid checks,
and set $\widetilde R_\ell=\operatorname{median}_{a\in\mathcal I_\ell}
\widehat R_a$ and $T_\ell=10^{-8}+10^{-4}\widetilde R_\ell$.
We risk convergence, we equire both
\begin{equation}\label{eq:risk-window-test}
 \max_{a\in\mathcal I_\ell}\widehat R_a
 -\min_{a\in\mathcal I_\ell}\widehat R_a\leq T_\ell,\qquad
 \max_{a,b\in\mathcal I_\ell}\frac1m\sum_j(f_a(Z_j)-f_b(Z_j))^2\leq T_\ell.
\end{equation}
Upon success we return the
lowest-monitoring-risk model in the final five-check window, breaking exact
ties in favor of the most recent model.
\subsection{Algorithm summary}
Algorithm~\ref{alg:alm-experiments} summarizes the solve for a fixed edge
set $W$. All its updates and acceptance tests are defined above. For a full
fit using constraint generation, this algorithm solves each working-set
problem; the wrapper applies the full-target checks before accepting the
full fit.

\begin{algorithm}[htbp]
\caption{Augmented Lagrangian with semismooth Newton subproblems}
\label{alg:alm-experiments}
\begin{algorithmic}[1]
\Require $X,Y,W,B,L,\mu_n$; data $\{(Z_j,f_0(Z_j))\}_{j=1}^m$;.
\State Initialize $w^0,\lambda^0,\nu^0,\sigma_0$ as above, or warm-start.
\For{$k=0,1,\ldots$}
 \State Approximately minimize $\psi_k$ using semismooth Newton and backtracking.
 \State If the inner accuracy test fails, stop with an unresolved status.
 \State Update $v^{k+1},\lambda^{k+1},\nu^{k+1}$ by \eqref{eq:alm-updates}.
 \State Update the feasible upper bound $U$ and dual lower bound $D$.
 \State Check the objective gap and risk convergence
 \State If accepted, return the selected predictor and its convergence label.
 \State Adjust $\sigma_{k+1}$ using $r_{k+1}$ and $r_k$.
\EndFor
\State If no acceptance test passes, record an unresolved fit.
\end{algorithmic}
\end{algorithm}

\section{Detailed experimental design}
\label{app:experiments}
We specifies our experimental procedures in this appendix

\subsection{Synthetic truth families}
\label{subsec:synthetic-details}
The primary domain is $\Omega=[-2,2]^d$, with independent uniform designs in dimensions $d\in\{2,3,4,5,6,8,10\}$. The benchmark amplitude is $A=8$; we use the symbol $A$ here to distinguish the prescribed truth scale from the optimization bound $B$. All four families below use the fixed common optimization bounds $A=8$ with $B = 10$ and $L=2B/\sqrt d=20/\sqrt d$ for each fixed dimension.

\subsubsection{The full and half quadratics}
The functions in \citet{BalazsGyorgySzepesvari2015} are
\begin{equation}\label{eq:balazs-benchmarks-appendix}
 f_{\rm Q}(x)=-A+\frac{A}{2d}\sum_{j=1}^d x_j^2,
 \qquad
 f_{\rm H}(x)=-A+\frac{A}{2d}\sum_{j=1}^d\max\{x_j,0\}^2.
\end{equation}

For $x\in[-2,2]^d$, both sums in \eqref{eq:balazs-benchmarks-appendix} lie between zero and $4d$, so both truths lie in $[-A,A]$. The full quadratic has
\[
 \nabla f_{\rm Q}(x)=\frac{A}{d}x,
 \qquad
 \nabla^2 f_{\rm Q}(x)=\frac{A}{d}I_d.
\]
It is therefore $(A/d)$-strongly convex, with Lipschitz constant at most $(A/d)2\sqrt d=2A/\sqrt d$ on the cube. Since $d$ is fixed in each asymptotic experiment, $m_0=A/d$ is a fixed positive curvature floor.

The scalar function $t\mapsto\max\{t,0\}^2$ is continuously differentiable, with derivative $2\max\{t,0\}$. Its derivative is nondecreasing, so the scalar function is convex. We sum coordinatewise to obtain convexity of $f_{\rm H}$ and
\[
 \nabla f_{\rm H}(x)=\frac{A}{d}x_+,
 \qquad
 \|\nabla f_{\rm H}(x)\|_2\le\frac{2A}{\sqrt d}.
\]
On $(-2,0)^d$ it is identically $-A$, ruling out any positive global strong-convexity constant. It has a change in curvature at coordinate hyperplanes, but its gradient is continuous; it should not be described as a max-affine slope kink.

\subsubsection{A fixed finite collection of planes}
For $M\in\{5,10,25,50\}$, we use
\[
 f_{\rm MA}(x)=\max_{1\le m\le M}\{a_m^\top x+b_m\}.
\]
We generate coefficients before drawing any regression data and save them. To actually generate the coefficients, we draw independent directions uniformly on the Euclidean unit sphere, sets $a_m$ to $2/\sqrt d$ times the direction, and draws $b_m$ uniformly on $[-1,1]$. Each coefficient realization is held fixed across the sample-size sweep and all training replications for that realization.

For every generated piece,
\[
 |a_m^\top x+b_m|
 \le\frac{2}{\sqrt d}(2\sqrt d)+1=5,
 \qquad
 \|a_m\|_2=\frac{2}{\sqrt d}.
\]
Thus $|f_{\rm MA}|\le5\le B$ and $\operatorname{Lip}(f_{\rm MA})\le2/\sqrt d\le L$. The maximum is convex, and it is not strongly convex because at least one full-dimensional affine region has nonempty interior. Due to the generation procedure, it is possible that some generated pieces we completely inactive, but we do not resolve this issue.

\subsubsection{The quadratic-interpolation hard family}
The random-design lower-bound construction of \citet[Lemma~3.2 and Appendix~B.6]{KurGaoGuntuboyinaSen2024} uses piecewise-affine interpolation of a quadratic, with approximation scale $k^{-2/d}$ and a complexity parameter of order $\sqrt n$ at fixed noise level.
For an integer $J\ge2$, we let
\[
 h_J=4/J,\qquad t_k=-2+kh_J\quad(k=0,\ldots,J),
\]
and define
\begin{equation}\label{eq:grid-chord-interpolant}
 q_J(t)=\max_{0\le k<J}
 \{(t_k+t_{k+1})t-t_kt_{k+1}\},
 \qquad -2\le t\le2.
\end{equation}
The $k$th affine term is the chord joining $(t_k,t_k^2)$ and $(t_{k+1},t_{k+1}^2)$. Consecutive slopes increase by $2h_J$, and the $k$th term is active on $[t_k,t_{k+1}]$. This follows directly by subtracting two adjacent terms: the $(k+1)$st term minus the $k$th term is $2h_J(t-t_{k+1})$. The maximum in \eqref{eq:grid-chord-interpolant} is therefore the ordinary convex linear interpolant of $t^2$.

We define
\begin{equation}\label{eq:hard-grid-truth}
 f_{{\rm hard},J}(x)
 =-A+\frac{A}{2d}\sum_{j=1}^d q_J(x_j),\qquad A=8.
\end{equation}
The sum of the coordinatewise maxima can be written as a single maximum of $J^d$ affine functions:
\[
 f_{{\rm hard},J}(x)
 =\max_{\mathbf k\in\{0,\ldots,J-1\}^d}
 \left\{-A+\frac{A}{2d}
 \sum_{j=1}^d\bigl[(t_{k_j}+t_{k_j+1})x_j-t_{k_j}t_{k_j+1}\bigr]\right\}.
\]
Each open grid cube has one distinct active affine function. The truth agrees with $f_{\rm Q}$ at all grid vertices and is affine throughout every grid cube, hence on any simplicial subdivision of those cubes. It is thus a concrete cube realization of the quadratic-interpolation construction, with explicitly verified convexity. It is an interpolating secant construction, not an arbitrary maximum of supporting tangent planes.

For $t\in[t_k,t_{k+1}]$, direct subtraction gives
\[
 q_J(t)-t^2=(t-t_k)(t_{k+1}-t).
\]
The product is nonnegative and at most $h_J^2/4$. We sum over coordinates to obtain
\begin{equation}\label{eq:hard-approximation-error}
 0\le f_{{\rm hard},J}(x)-f_{\rm Q}(x)
 \le\frac{Ah_J^2}{8}=h_J^2=\frac{16}{J^2}.
\end{equation}
The interpolation values lie in $[0,4]$ and $|t_k+t_{k+1}|\le4$, so $f_{{\rm hard},J}\in[-A,A]$ and every affine slope has norm at most $2A/\sqrt d$. These bounds hold uniformly in $J$. Since the function is affine on nonempty open cells, it has no positive global curvature floor.

For the principal high-dimensional stress test, we use
\[
 J_n=\max\{2,\lceil n^{1/(2d)}\rceil\},\qquad
 f_{0,n}=f_{{\rm hard},J_n},\qquad M_n=J_n^d.
\]
Then $M_n\asymp\sqrt n$, the sup-norm approximation error to $f_{\rm Q}$ is $O(n^{-1/d})$, and its squared population discrepancy from $f_{\rm Q}$ is $O(n^{-2/d})$. Integer rounding makes $J_n$ piecewise constant at accessible sample sizes, especially in large dimension, so record the actual pair $(J_n,M_n)$ in every run. A fixed-grid experiment with $J$ held constant is a distinct, fixed-truth control. Both designs choose the truth independently of all regression observations.

This changing-truth family is valid for our uniform upper theorem: for every $n$, $f_{0,n}$ belongs to the same fixed class $\mathcal F_{B,L}(\Omega)$. Its observed risk curve is not a pointwise asymptotic curve for one fixed truth. The lower-bound result in the cited paper concerns its specified LSEs and appropriately chosen construction constants; we do not transfer that lower bound to our regularized programs or claim that every finite grid is least favorable. The experiment probes the mechanism of that construction and the full/KNN comparison.

\subsection{Sample sizes, regularization, and neighborhood grids}
\label{subsec:threshold-grid}
For each fixed dimension, we use independent training replications at
\begin{equation}\label{eq:study-sample-grid}
 n = \{250, 300,400,500, 600, 700, 800, 900,1000, 1500,2000,2500,3000\}
\end{equation}
with $30$ replications as the research configuration. Matching our assumption from \ref{ass:regularization}, we set our penalty to
\[
 \mu_n=c_\mu r_{n,d}/L^2,\quad c_\mu=0.2,
\]
with $c_\mu\in\{0.05,0.2,1\}$ as separately reported sensitivity settings. The values $B$, $L$, $d$, and $c_\mu$ are held fixed within a sample-size sweep. At each $n$, the full and every KNN problem use exactly the same $\mu_n$; it is not retuned as $K$ changes. Fixed-plane coefficients are generated independently of the training data and reused for all $n$ and all replications. The hard family instead uses its explicitly prescribed $J_n$ and is labelled as a changing-truth experiment.

We denote $s(n,d)$ denote the sufficient neighborhood schedule found in Theorem \ref{thm:general-summary} and \ref{thm:strong-summary} before multiplying
by the constant $C_K$. 
The full quadratic uses $s_{\rm sc}$; the other families use $s_{\rm gen}$.
The study grid is the union of $n-1$, $\lceil c \times s(n,d)\rceil$ for
$c\in\{0.05,0.1,0.25,0.5,1,2\}$, and $\lceil n^\alpha\rceil$ for
$\alpha\in\{0.4,0.6,0.8,0.85,0.9,0.95\}$.

\subsection{Empirical risk-transition experiment}
\label{subsec:empirical-threshold}
For each truth specification, fixed $n,d$, and training replicate $b\in\{1,\ldots,R\}$, we use the same training observations for the full estimator and every KNN fit. We evaluate the numerical predictors $\widetilde f_{b,K}$ and $\widetilde f_{b,F}$ at an independent common test set $Z_{b,1},\ldots,Z_{b,M_{\rm test}}$, using the known noiseless truth. For a completed fit, the final test risk is
\begin{equation}\label{eq:test-risk}
 L_{b,K}=\frac1{M_{\rm test}}\sum_{j=1}^{M_{\rm test}}
       (\widehat f_{b,K}(Z_{b,j})-f_0(Z_{b,j}))^2.
\end{equation}
The same test points are used across $K$ within replicate $b$. We then compare the ratio of the KNN and full test risk against $K$.
\end{document}